\documentclass{article}

\usepackage{PRIMEarxiv}

\usepackage[utf8]{inputenc}
\usepackage[T1]{fontenc}
\usepackage[numbers,sort&compress]{natbib}
\usepackage{hyperref}
\usepackage{url}
\usepackage{booktabs}
\usepackage{amsmath}
\usepackage{amssymb}
\usepackage{amsfonts}
\usepackage{mathtools}
\usepackage{nicefrac}
\usepackage{array}
\usepackage{microtype}
\usepackage{xcolor}
\usepackage{graphicx}
\usepackage{algorithm}
\usepackage{algorithmicx}
\usepackage{algpseudocode}
\usepackage{placeins}
\usepackage{float}
\usepackage{fancyhdr}

\hypersetup{
  colorlinks=true,
  linkcolor=black,
  citecolor=black,
  urlcolor=blue,
  breaklinks=true,
  pdftitle={DISEIL: Demonstration Distillation for Sample-Efficient Imitation Learning},
  pdfauthor={Suyog Khanal, Arun Kumar A V, Santu Rana}
}

\title{DISEIL: Demonstration Distillation for\\Sample-Efficient Imitation Learning}

\author{
  Suyog Khanal \quad Arun Kumar A V \quad Santu Rana \\
  \normalfont Deakin Applied Artificial Intelligence Initiative \\
  \normalfont Geelong, Australia
}

\begin{document}

\maketitle

% !TeX root = ../main.tex
\begin{abstract}
A robot that can be taught a new task from a handful of demonstrations has to work out for
itself what it still cannot do, and then ask for exactly that. Interactive imitation learning
takes a step in that direction by letting a policy practice on its own and calling an expert
when it goes wrong. Existing methods decide when to interrupt the learner. A further 2 decisions are left to whichever
episode happened to trigger the interruption: which failure to correct, and where the
demonstration should start. This paper is a first attempt at making both
of them deliberately. \textbf{DISEIL} (\textbf{D}emonstration d\textbf{I}stillation for
\textbf{S}ample-\textbf{E}fficient \textbf{I}mitation \textbf{L}earning) marks each failed
episode at the step where the policy first becomes unreliable, represents that moment with a
geometric descriptor, and groups the failures into recurring failure modes. A vision-language
model and a language model read the selected mode and write a request for the next
demonstration, and a store of task constraints checks that the request can be carried out
before any expert time is spent. No model produces a robot action. Across 5 simulated tasks under state and image observations, changing only what the expert is
asked for gives the highest mean held-out success rate in all 10 settings, with a tie in 1, and the margin is
widest at the smallest budget we tested. The scope is narrow: a single round of practice at a time, in simulation, with experts that are
mostly scripted. The longer-term aim is a learner that also tracks what its demonstration set
already covers, and that asks a human teacher for the missing behavior in proportion to the
effort each request costs them.
\end{abstract}


\section{Introduction}

\begin{figure}[t]
\centering
\includegraphics[width=0.40\linewidth]{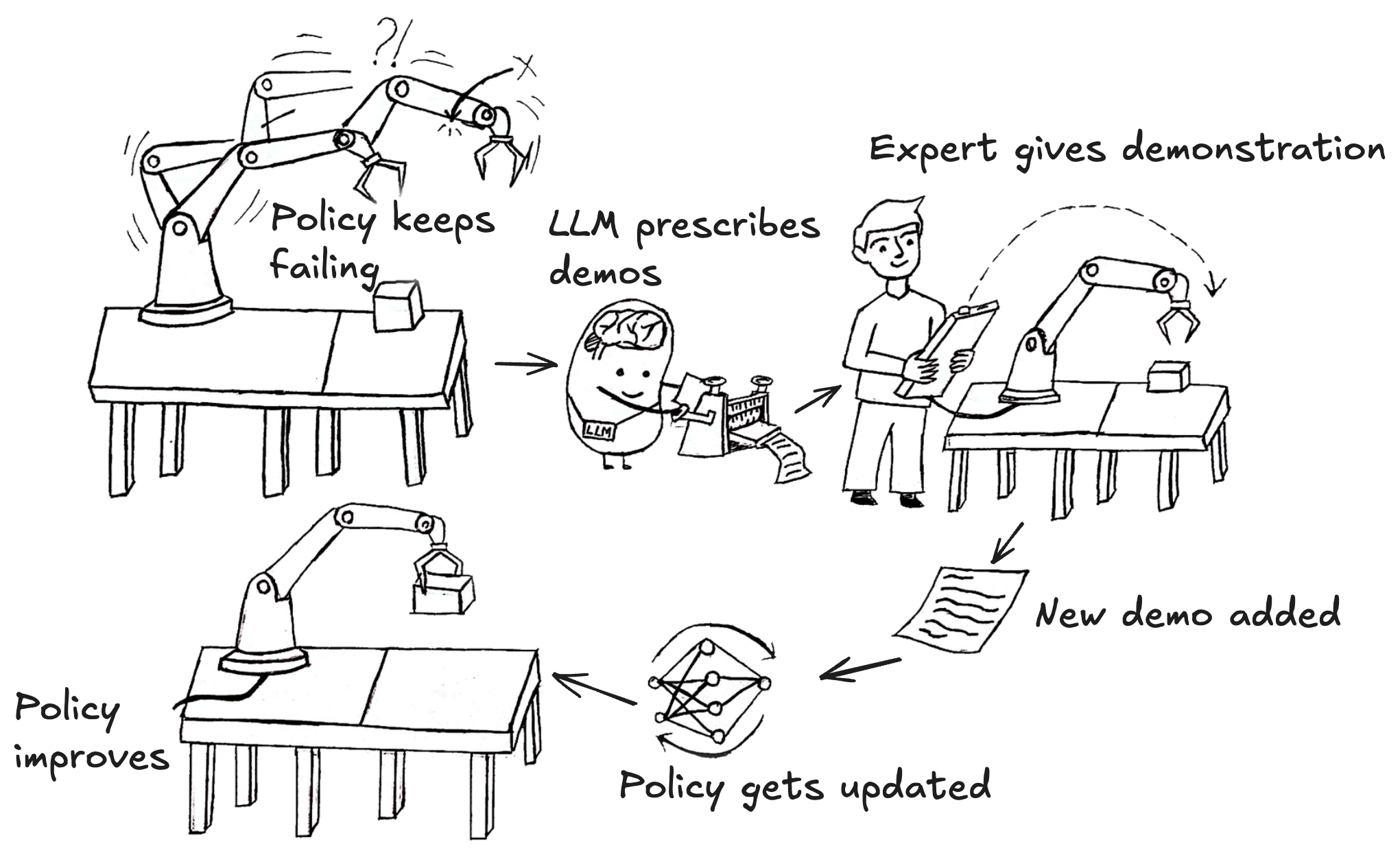}
\caption{Existing methods correct the single failure that triggered the query. DISEIL looks for
patterns across failures and asks for a demonstration that starts at a chosen configuration.}
\label{fig:teaser}
\end{figure}

Behavior cloning trains a policy on the actions an expert took in the states the expert visited
\citep{pomerleau1988alvinn,bain1995cloning}. Small mistakes then take the learner into states it
has never seen, where further mistakes pile up. This is covariate shift
\citep{shimodaira2000covariate,ross2011dagger}. Interactive imitation learning reduces the
problem by letting the expert correct the learner in the states it actually reaches
\citep{ross2011dagger,celemin2022iil}. What it cannot do is make the expert cheap.
Demonstrations are produced one at a time by a person or an oracle, and they are usually the
scarcest resource \citep{mandlekar2018roboturk,khazatsky2024droid}. How well a policy performs
depends on what is in the demonstration set and not only on how large it is
\citep{lin2024datascaling}. So once the budget is fixed, the question that remains is what each
demonstration should contain (Figure~\ref{fig:teaser}).

Getting a demonstration takes 3 decisions: when to call the expert, which failed episode to
correct, and where the demonstration should start. The DAgger family answers the first one, and
its members differ mainly in the signal they watch. Some measure disagreement between the
learner's action and the expert's \citep{zhang2017safedagger}. Others measure how much the
prediction varies under dropout \citep{menda2017dropoutdagger} or across an ensemble
\citep{menda2019ensembledagger}, combine novelty with risk \citep{hoque2021thriftydagger}, or
read a diffusion policy's per-step denoising loss \citep{lee2025diffdagger,chi2023diffusionpolicy}.
The other 2 decisions are usually not discussed at all. They default to the episode that
triggered the query and the state that episode had already reached.

Those defaults do no harm when the budget is large, because repeated corrections will eventually
cover every weakness. They do harm when the budget is small. A score attached to a single state says nothing about whether 2 failures come from the same
underlying mistake, so the budget can be spent again and again on a single common weakness while
others are never touched. And because the
expert is only called after the signal fires, a demonstration that starts at the triggering state
may teach the learner how to recover from a mistake whose effects have already built up, rather
than how to avoid making it.

\paragraph{Our position.}
We argue that a demonstration should be aimed at a weakness the policy actually has, rather than treated as just another unit of supervision. In practice this means writing down the round's failures
explicitly, grouping them into recurring \emph{failure modes}, choosing which mode to supervise,
and saying where the demonstration should start. All of that happens before the expert is called.
DISEIL is one way of doing this, and it is built so that the query rule, the policy, the training
objective and the evaluation protocol all stay the same. Only the content of the request changes.

\paragraph{Why part of the loop is neural and part is symbolic.}
The work is split according to what current models do well and what they do badly.
Vision-language models can describe what a robot did and name a plausible cause when they are
given structured evidence \citep{liu2023reflect,duan2025aha}. They are far less reliable at
judging distances and positions from images \citep{chen2024spatialvlm,fu2024blink}, and a
configuration they invent is useless if the robot cannot physically reach it. DISEIL therefore
never asks a model to work out the geometry of the failures. The failures are grouped using an
explicit geometric descriptor of the configuration at the step where each failure begins. The
models supply the reading of what went wrong and write the request. A store of task constraints,
together with a few discrete feasibility checks, then decides whether the request may be sent to
the expert. The models run only while a demonstration is being chosen. They produce no actions
and are not part of the control loop at run time.

\paragraph{What this paper reports.}
Section~\ref{sec:method} states the acquisition problem so that its 3 decisions are separate and explicit: when the expert is called, which recurring failure mode receives the next
demonstration, and which configuration that demonstration starts from. It then describes DISEIL as a 4-stage loop over that problem. Section~\ref{sec:study} reports a
study across 5 tasks and 2 observation types in which the way demonstrations are chosen is the
only thing that changes. Given 20 demonstrations, DISEIL obtains the highest mean held-out success
rate in all 10 settings. A first pass at removing components one at a time associates most of that
gain with grouping the failures and allocating across the groups, rather than with the language
models. That decomposition rests on 1 arm per setting on 3 of the 10 settings, so we read it as an initial indication that motivates refining the allocation stage and running a broader
ablation, not as a settled account. Section~\ref{sec:scope} states the limits of the evidence.
The geometric descriptor is designed by hand and reads privileged simulator state, every task is
simulated, most of the experts are scripted, and the budget counts demonstrations rather than
expert effort.

% !TeX root = ../main.tex
\section{Background}
\label{sec:related}

\paragraph{Interactive imitation learning and when to call the expert.}
Dataset aggregation reduces covariate shift by retraining the policy on the states the learner
visits \citep{shimodaira2000covariate,ross2010reductions,ross2011dagger}, and every baseline we
compare against follows that loop. They differ in the score that decides when control passes to
the expert
\citep{zhang2017safedagger,menda2017dropoutdagger,menda2019ensembledagger,hoque2021thriftydagger,lee2025diffdagger}.
Some variants leave that decision to a human operator \citep{kelly2019hgdagger} or change when
control is handed back \citep{hoque2021lazydagger}. Intervention-weighted methods put extra
weight on expert-labeled corrective states during training
\citep{mandlekar2020iwr,liu2023sirius}, which changes how the collected data are used rather than
which failure gets corrected next. In all of this work, the episode that triggered the query and
the state it had reached still decide what gets supervised and where the demonstration begins.

\paragraph{Choosing existing data against specifying new data.}
Active learning ranks candidates by uncertainty or coverage
\citep{settles2009active,houlsby2011bald}. Core-set selection uses greedy $k$-center
\citep{sener2018coreset}, BADGE uses gradient embeddings \citep{ash2020badge}, sub-trajectory
retrieval ranks segments of a corpus that already exists \citep{memmel2025strap}, and dataset
distillation compresses a training set after it has been collected \citep{cazenavette2022mtt}.
All of these pick from data that is already there. Demonstration distillation runs before the
demonstration exists. Evidence from several failed episodes becomes a specification for
supervision that has not been collected yet, and the expert then produces the trajectory that was
asked for.

\paragraph{Neural reading with symbolic checking.}
Language models have been used for robot planning, code generation and reward design
\citep{ahn2022saycan,liang2023codeaspolicies,ma2024eureka}, and for explaining failures from
video and trajectory evidence \citep{liu2023reflect,duan2025aha}. DISEIL uses them only in the
second role. The store of workspace, spawn, reachability and success constraints that every
proposal is checked against is a robot knowledge base in the sense of
\citet{tenorth2013knowrob}, not a text retrieval index \citep{lewis2020rag,edge2024graphrag}, and
the feasibility checks it drives are ordinary discrete search
\citep{hart1968astar,cormen2022algorithms}. Using an external planner and revising a proposal
after a checker rejects it are both established ideas
\citep{liu2023llmp,chen2024autotamp}. What is new here is applying them to the design of the
demonstration that gets collected.

% !TeX root = ../main.tex
\section{Demonstration distillation}
\label{sec:method}

\paragraph{The problem.}
A policy $f_\theta$ maps observations to actions. For every state and action it encounters it
also reports a per-step loss $\ell^{(i)}_t=\mathcal{L}(f_\theta, s^{(i)}_t, a^{(i)}_t)$, where
$i$ indexes episodes and $t$ indexes time steps. For a diffusion policy $\mathcal{L}$ is the
denoising loss; for a discrete policy it is the negative log-likelihood of the action that was
taken \citep{chi2023diffusionpolicy}. The policy is first trained on an initial set
$\mathcal{D}_0$ by behavior cloning. Round $r$ then runs the policy from a fresh set of starting
configurations drawn from the task's reset distribution. The episodes it fails, together with
their loss traces, make up the failure set $\mathcal{F}_r$, and we write $N=|\mathcal{F}_r|$ for
its size. The round ends by collecting $D$ demonstrations from the expert $\pi^\star$ and
retraining $\theta$ \citep{ross2011dagger}. The whole loop stops when
$|\mathcal{D}_r|-|\mathcal{D}_0|=B$. Success is measured on a fixed set of held-out starting
states that every method shares and that is never used for collection.

Under a fixed budget $B$, an acquisition decision is the triple
$A_r=(t^\star, C_{\mathrm{tgt}}, \xi)$. Here $t^\star$ is when the query fires,
$C_{\mathrm{tgt}}$ is the failure mode that gets supervised, and $\xi$ is the configuration the
demonstration starts from. Query-gated methods set $t^\star$
\citep{hoque2021thriftydagger,lee2025diffdagger} and let the other 2 fall to the triggering episode and its triggering state. We keep the query rule as it is and work on the rest: choosing
$C_{\mathrm{tgt}}\subseteq\mathcal{F}_r$ and $\xi$ so that a single demonstration covers as much
of the observed failure distribution as it can.

\begin{figure}[t]
\centering
\includegraphics[width=0.96\linewidth]{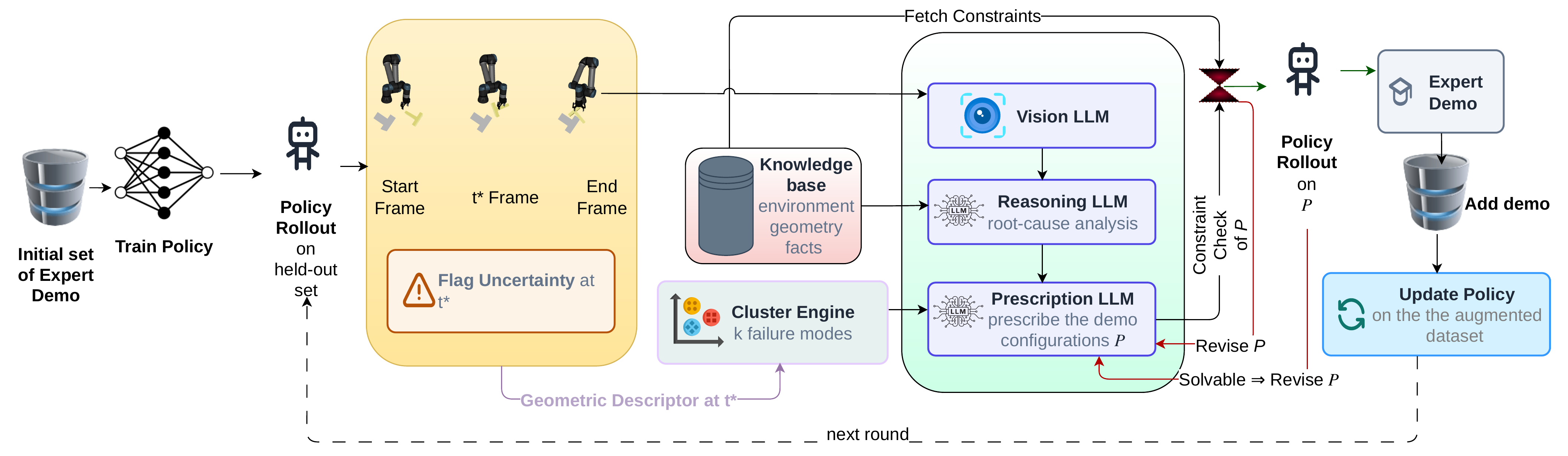}
\caption{A single acquisition round. The query gate marks each failure at $t^\star$. A geometric
descriptor of each failure groups them into modes and picks the target. Neural annotations
describe the chosen mode. The constraint store decides whether the resulting request may be
sent to the expert.}
\label{fig:architecture}
\end{figure}

DISEIL runs 4 stages per round (Figure~\ref{fig:architecture}). \textbf{Perceive} turns each
failure into a geometric descriptor and a written description. \textbf{Partition} groups those
descriptors into modes. \textbf{Prioritize} picks the mode to correct. \textbf{Prescribe} says where the next
demonstration should start. A vision-language model (VLM) \citep{bai2025qwen3vl} describes
rendered observations and a text-only prescription model (PM) \citep{yang2025qwen3} assigns
labels and writes the request. Neither takes part in the grouping.

\paragraph{Perceive.}
Failure $i$ is marked at the first step whose loss stays above a threshold $\eta$ for $K$ steps
in a row,
\begin{equation}
\label{eq:flag}
t^\star_i=\min\big\{t:\ \ell^{(i)}_u>\eta\ \ \forall u\in[t,\,t+K-1]\big\},
\end{equation}
and at $\arg\max_t \ell^{(i)}_t$ if no such step exists. The threshold is a quantile of the
policy's own training-loss distribution and is recomputed every time the policy is retrained.
This is the Diff-DAgger query rule, used without changes \citep{lee2025diffdagger}. Marking the
first sustained crossing puts the anchor where the failure starts, rather than at a later state
where its effects have already built up. The robot and object state at $t^\star_i$ is then written as
a geometric descriptor,
\begin{equation}
\label{eq:desc}
\phi_i=\big[p_{x,i},\,p_{y,i},\,\sin\psi_i,\,\cos\psi_i,\,\rho_i,\,\delta_i\big],
\qquad
\rho_i=t^\star_i/T_i,
\qquad
\delta_i=\big\|p^{\mathrm{ee}}_i-p^{\mathrm{obj}}_i\big\|_2,
\end{equation}
where $(p_{x,i},p_{y,i})$ and $\psi_i$ are the position and yaw of the object the task is about,
$\rho_i$ is how far through the episode of length $T_i$ the failure started, and $\delta_i$ is
the distance from the end-effector to the object. Writing yaw as $(\sin\psi_i,\cos\psi_i)$ avoids
a false jump between $-\pi$ and $+\pi$, which are the same orientation. We keep the descriptor
low-dimensional because a round only holds a few dozen failures, and every task fills the same
slots, so the grouping works the same way on every task and observation type. In parallel, the VLM
summarizes frames from the start of the episode, the marked step and the end into a short spatial
description. The PM turns that into a root cause and a trajectory phase, both drawn from a fixed
vocabulary held in the task knowledge base $\mathcal{K}$, which also stores the workspace and
reachability constraints. The 2 streams stay separate until Prescribe.

\paragraph{Partition.}
When $N\geq4$, the descriptors are standardized one dimension at a time and the number of modes
is chosen by silhouette score, which measures how well separated a grouping is:
\begin{equation}
\label{eq:kstar}
\tilde{X}_i=\frac{\phi_i-\mu_\phi}{\sigma_\phi},
\quad
k_{\max}=\min\{6,N-1\},
\quad
k^\star=\operatorname*{argmax}_{2\leq k\leq k_{\max}}\operatorname{sil}(k),
\quad
\{C_1,\ldots,C_{k^\star}\}=\mathcal{C}(\tilde{X},k^\star).
\end{equation}
We use agglomerative clustering for $\mathcal{C}$ \citep{ward1963hierarchical} and pick
$k^\star$ with the standard silhouette criterion
\citep{rousseeuw1987silhouette,pedregosa2011sklearn}. Nothing rests on that particular choice.
When $N<4$ each failure becomes its own mode, because silhouette scores are not reliable on so
few points. Each cluster stands in for a \emph{failure mode}. It stores how many failures it
holds, its centroid in task space, its mean peak loss
$\bar{L}_C=|C|^{-1}\sum_{i\in C}\max_t\ell^{(i)}_t$ as a measure of how serious it is, and a
representative $\mathrm{rep}(C)$, the member closest to the cluster mean. The largest cluster is
$C^\star$, with ties broken by $\bar{L}_C$, and each cluster is labeled with the root cause most
of its members share rather than with a number.

\paragraph{Prioritize.}
Always taking the largest mode would keep returning to modes that look alike from one round to
the next. DISEIL therefore combines how often a mode occurs with a per-task cluster memory
$\mathcal{M}$ that lowers the score of modes similar to ones already corrected. Writing
$\bar{L}^{\mathcal{M}}_C$ for the severity after that adjustment, the target among the
near-largest modes is
\begin{equation}
\label{eq:target}
C_{\mathrm{tgt}}=\operatorname*{argmax}_{C:\,|C|\geq|C^\star|-1}\bar{L}^{\mathcal{M}}_C .
\end{equation}
Size favors corrections that apply to several episodes, severity favors the most serious
weakness, and the memory discourages fixing the same thing twice in a row. The PM then receives
at most $\kappa$ failures from $C_{\mathrm{tgt}}$: first $\mathrm{rep}(C_{\mathrm{tgt}})$, then
the member with the highest peak loss, and any remaining slots are filled by farthest-point
sampling \citep{eldar1997fps}. The resulting context set $S$ contains a central example, a severe example, and geometrically
diverse evidence from the selected mode.

\paragraph{Prescribe.}
Given the geometry of the target mode, the annotated failures in $S$ and the constraints in
$\mathcal{K}$, the PM returns either a targeted correction or a bridging placement. A \emph{targeted correction}
puts a cited episode back at its marked state and asks the expert to carry on from there. A \emph{bridging
placement} builds a new starting configuration from 2 or more related failures, and that
configuration need not coincide with any state that was recorded. Reverse curriculum learning
\citep{florensa2017reversecurriculum} generates reset states for a policy; we use the same
construction to place the start of an expert demonstration. On attempt $j$ the command
$\mathrm{PM}(C_{\mathrm{tgt}},S,\mathcal{K},a^{(j-1)})$ is decoded into a concrete configuration
$\xi^{(j)}$, which is accepted only if
\begin{equation}
\label{eq:verify}
A_\theta(\xi)=
\underbrace{
\mathbf{1}\big[\xi\in\mathcal{W}_{\mathcal{K}}\big]\wedge
\mathbf{1}\big[\mathrm{reachable}_{\mathcal{K}}(\xi)\big]\wedge
\mathbf{1}\big[\mathrm{valid\text{-}path}_{\mathcal{K}}(\xi)\big]
}_{V_{\mathcal{K}}(\xi),\ \text{the symbolic check}}
\wedge\ \mathbf{1}\big[f_\theta\text{ does not solve }\xi\big] .
\end{equation}
On the grid task, whether a valid path exists is decided by A\textsuperscript{$\star$} and
breadth-first search \citep{hart1968astar,cormen2022algorithms}. A rejected attempt records why
it was rejected in $a^{(j)}$, either the constraint it violated or the fact that the policy
already solves it, and that reason is passed back to the PM for the next attempt, starting from
$a^{(0)}=\emptyset$. Rejected proposals cost no expert time. After $J_{\max}$ failed attempts
DISEIL falls back to the nearest untried marked failure, so every round still produces a valid
demonstration without wasting a query. Once a proposal is accepted the expert demonstrates from
$\xi$, the corrected mode is added to the memory, and the policy is retrained on the fixed
schedule. The loop is written out as pseudocode in the supplement.

% !TeX root = ../main.tex
\section{A controlled study}
\label{sec:study}

% Table 1: final held-out success rate
\begin{table}[t]
\centering
\caption{Held-out success rate (\%) after $B=20$ expert demonstrations, as mean $\pm$ standard
error over 9 GridWorld seeds and 5 robot-task seeds. $|\mathcal{D}_0|$ is the number of
demonstrations the policy started with and Init SR is its success rate before any were added.
Bold marks the highest mean in each row, with a tie bolded on both sides. A dash means the method
does not apply to that setting or was not run there.}
\label{tab:main}
\footnotesize
\setlength{\tabcolsep}{4.5pt}
\begin{tabular}{llcc ccccccc}
\toprule
Task & Obs & $|\mathcal{D}_0|$ & Init SR & Safe & Dropout & Ensemble &
Thrifty & Stagger & Diff-DAgger & DISEIL (ours) \\
\midrule
GridWorld & state & 20 & 48.9 &
85.3$\pm$0.9 & 84.9$\pm$0.8 & 86.2$\pm$0.7 & 86.8$\pm$0.7 &
85.7$\pm$0.5 & -- & \textbf{92.4$\pm$0.4} \\
GridWorld & image & 20 & 47.0 &
88.8$\pm$0.9 & 88.4$\pm$0.7 & 88.8$\pm$0.9 & 88.7$\pm$0.6 &
89.1$\pm$0.8 & -- & \textbf{91.3$\pm$0.6} \\
Push-T & state & 20 & 46.2 &
82.0$\pm$3.0 & 84.8$\pm$2.7 & 85.9$\pm$2.6 & 83.2$\pm$3.2 &
-- & 94.1$\pm$2.0 & \textbf{96.1$\pm$1.6} \\
Push-T & image & 20 & 43.3 &
78.1$\pm$3.5 & 82.1$\pm$3.1 & 83.2$\pm$3.0 & 79.3$\pm$3.6 &
-- & 89.0$\pm$2.1 & \textbf{92.6$\pm$2.2} \\
Lift & state & 8 & 67.2 &
99.2$\pm$0.7 & 99.2$\pm$0.4 & 99.2$\pm$0.4 &
\textbf{100.0$\pm$0.0} & -- & 99.2$\pm$0.4 &
\textbf{100.0$\pm$0.0} \\
Lift & image & 8 & 66.4 &
99.6$\pm$0.4 & 97.2$\pm$1.6 & 98.8$\pm$0.7 & 99.6$\pm$0.4 &
-- & 99.6$\pm$0.4 & \textbf{100.0$\pm$0.0} \\
Wipe & state & 12 & 47.7 &
88.0$\pm$1.1 & 88.6$\pm$1.8 & 86.8$\pm$1.9 & 89.0$\pm$1.1 &
-- & 90.4$\pm$2.7 & \textbf{93.1$\pm$1.3} \\
Wipe & image & 12 & 45.2 &
69.6$\pm$2.4 & 83.2$\pm$3.0 & 84.4$\pm$3.2 & 69.2$\pm$4.0 &
-- & 88.6$\pm$1.4 & \textbf{92.3$\pm$1.4} \\
Door & state & 4 & 56.8 &
91.8$\pm$2.1 & 92.5$\pm$1.2 & 88.8$\pm$3.1 & 89.6$\pm$1.7 &
-- & 93.2$\pm$1.9 & \textbf{96.6$\pm$1.9} \\
Door & image & 4 & 43.1 &
82.4$\pm$1.4 & 81.8$\pm$1.5 & 83.0$\pm$4.9 & 82.8$\pm$1.2 &
-- & 84.2$\pm$1.6 & \textbf{88.6$\pm$1.5} \\
\bottomrule
\end{tabular}
\end{table}

The study is set up so that the way demonstrations are chosen is the only thing that changes.
Within a setting, every method uses the same policy, the same starting demonstrations, the same
expert, the same retraining schedule, the same budget and the same held-out evaluation set.

\paragraph{Setting.}
We evaluate 5 tasks, each under state and image observations, which gives 10 settings. GridWorld
is a $5\times5$ grid containing 3 obstacles. Push-T is the planar pushing task of
\citet{florence2021implicitbc}, run in ManiSkill3 \citep{tao2024maniskill3}. Lift, Wipe and Door
are UR5/UR5e manipulation tasks in the RoboSuite simulator \citep{zhu2020robosuite}. All 5 tasks are simulated; no experiment in this paper runs on physical hardware. GridWorld uses an MLP
on state and a CNN on images. The robot tasks use diffusion policies
\citep{chi2023diffusionpolicy} with R3M encoding the images \citep{nair2022r3m}. R3M supplies
features to the policy only; the grouping uses the geometric descriptor of Eq.~\eqref{eq:desc}.
The expert is a person on GridWorld, a PPO policy on Push-T \citep{schulman2017ppo}, and a
scripted or motion-planned oracle on Lift, Wipe and Door. Qwen3-VL-32B does the visual analysis
and Qwen3-32B writes the requests. Every method gets $B=20$ demonstrations at $D=1$ per round, and the policy is retrained from
scratch after each one. We report held-out success on starting
states that were never used for collection, as mean $\pm$ standard error over 9 GridWorld seeds
and 5 robot-task seeds. A behavior-cloning sweep sets $|\mathcal{D}_0|$ per task so that success
before any demonstrations are added sits near 50\%. At that level the policy fails often enough
for recurring structure to appear in the failures, and it retains enough headroom for the choice
of demonstration to affect the final result. The comparisons are the DAgger-family gates
\citep{ross2011dagger,zhang2017safedagger,menda2017dropoutdagger,menda2019ensembledagger,hoque2021thriftydagger,lee2025diffdagger}.
Diff-DAgger uses the same denoising loss DISEIL marks failures with, so it applies only to the
robot tasks. STAGGER \citep{li2025interactivehybridil} appears only on GridWorld, because it asks
the expert for actions at single states rather than for whole trajectories. DISEIL computes its
geometric descriptors from privileged robot and object state. That state is used only for
grouping and is never given to the policy being trained. The supplement gives the tasks, experts,
baselines and hyperparameters in full.

\begin{figure}[t]
\centering
\includegraphics[width=\linewidth]{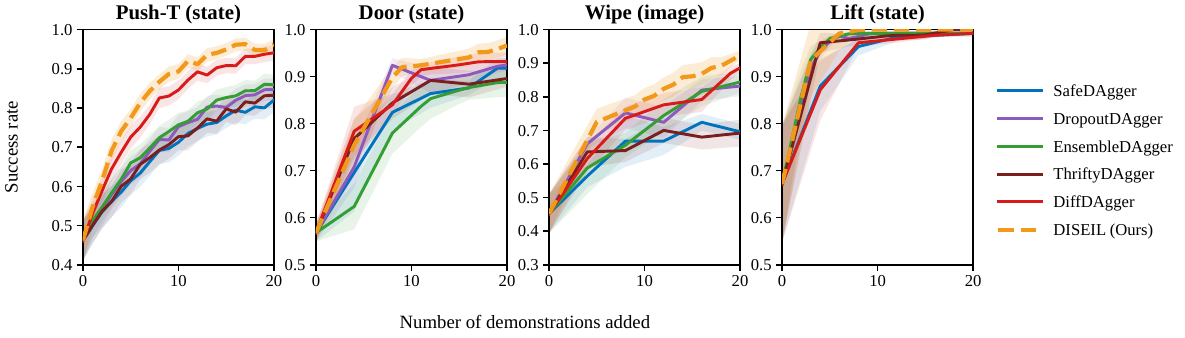}
\caption{Held-out success rate against the number of demonstrations collected, as mean $\pm$
standard error over runs with slightly different starting policies. DISEIL separates from the strongest Push-T baselines after about 5 demonstrations and holds the advantage for the rest of the budget.}
\label{fig:curves}
\end{figure}

\paragraph{Final success rate.}
DISEIL obtains the highest mean final success rate in all 10 settings (Table~\ref{tab:main}),
with a tie on Lift under state observations. Its margin over the strongest comparison in each
setting has a mean of 2.80 percentage points and ranges from 0.0 to 5.6. Which method is second
depends on the task: Diff-DAgger is the strongest comparison on Push-T, Wipe and Door, while the
simpler gates remain competitive on GridWorld. The improvement is therefore consistent in sign across settings rather than large in any single setting. On GridWorld with images, for example,
STAGGER reaches 89.1\% and the query-gated methods reach 88.4\% to 88.8\%, against DISEIL's
91.3\%. Figure~\ref{fig:curves} shows the separation appearing early in the budget and persisting. On Wipe with images,
both DISEIL and Diff-DAgger are still improving when the budget runs out. On Lift, DISEIL reaches 100\% after 9 demonstrations with state observations, 8 earlier than
ThriftyDAgger, and after 17 with images.

\begin{figure}[t]
\centering
\includegraphics[width=0.70\linewidth]{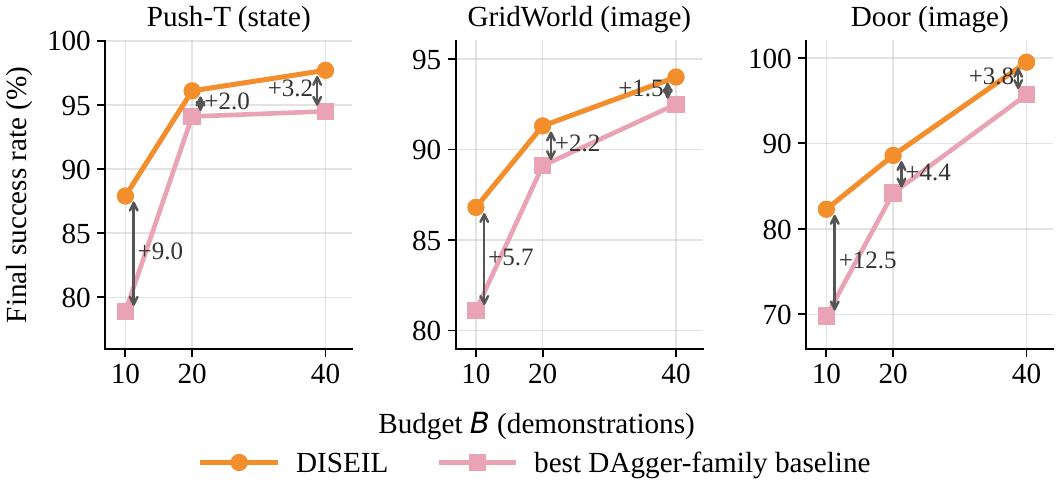}
\caption{Final success rate at budgets $B\in\{10,20,40\}$ on GridWorld with images, Push-T with state
observations and Door with images. Labels give DISEIL's margin over the strongest baseline in that
setting. The margin is widest at the smallest budget.}
\label{fig:budget}
\end{figure}

\paragraph{Effect of the size of the budget.}
We repeat the comparison at 3 budgets, $B=10$, $B=20$ and $B=40$, on 3 of the 10 settings: GridWorld with images, Push-T with state observations, and Door with images
(Figure~\ref{fig:budget}). Averaged over those 3 settings, DISEIL's margin over the strongest
baseline is 9.07 points at $B=10$, 2.87 points at $B=20$ and 2.83 points at $B=40$. DISEIL's own
success rate rises with the budget in all 3 settings; the margin narrows because the
baselines improve faster over the same range. The result does not show that DISEIL matches the
baselines on half the demonstrations. It shows that the allocation decision has most influence when the budget is small, which is the
regime in which expert supervision is the binding constraint.

% Table 2: per-demonstration goodness (caption above the table, per NeurIPS convention)
\begin{table}[!tb]
\centering
\caption{Per-demonstration goodness (the policy's per-step loss on a demonstration
before retraining), reported as mean $\pm$ standard error. Diff-DAgger queries using
the same per-step loss.}
\label{tab:infogain}
\footnotesize
\setlength{\tabcolsep}{4pt}
\begin{tabular}{llcc@{\hskip 1.6em}llcc}
\toprule
Task & Obs & Diff-DAgger & DISEIL & Task & Obs & Diff-DAgger & DISEIL \\
\midrule
Push-T & state & 1.57$\pm$0.49 & \textbf{2.81$\pm$0.93} & Wipe & state & 1.43$\pm$0.36 & \textbf{2.91$\pm$0.90} \\
Push-T & image & 1.80$\pm$0.49 & \textbf{2.82$\pm$0.77} & Wipe & image & 1.95$\pm$0.52 & \textbf{3.62$\pm$0.98} \\
Lift   & state & 1.61$\pm$0.50 & \textbf{2.64$\pm$0.74} & Door & state & 1.84$\pm$0.50 & \textbf{3.43$\pm$0.95} \\
Lift   & image & 1.36$\pm$0.38 & \textbf{2.93$\pm$0.75} & Door & image & 1.58$\pm$0.41 & \textbf{3.00$\pm$0.89} \\
\bottomrule
\end{tabular}
\end{table}

\paragraph{Per-demonstration goodness.}
We summarize the \emph{goodness} of an acquired demonstration by the policy's
per-step loss on it before retraining (Table~\ref{tab:infogain}); a higher value
means the demonstration lies where the current policy is most wrong, and so has
the most to teach. Diff-DAgger is the closest comparator because it queries using
the same loss, yet DISEIL acquires demonstrations of higher goodness in all 8 robot settings. Supplementary results show the same ordering against the other
baselines. Because goodness is measured before the demonstration is added to
training, it isolates how novel each demonstration is to the current policy
rather than its eventual effect after retraining. DISEIL also uses peak loss when
prioritizing modes (Eq.~\eqref{eq:target}), so the metric does not uniquely favor
Diff-DAgger. The result shows that failure-mode allocation preserves individual
demonstration informativeness while reducing repeated correction of similar
failures.

% Figure 5: Push-T failure-mode visualization
\begin{figure}[!tb]
\centering
\includegraphics[
    width=\linewidth,
    height=0.30\textheight,
    keepaspectratio
]{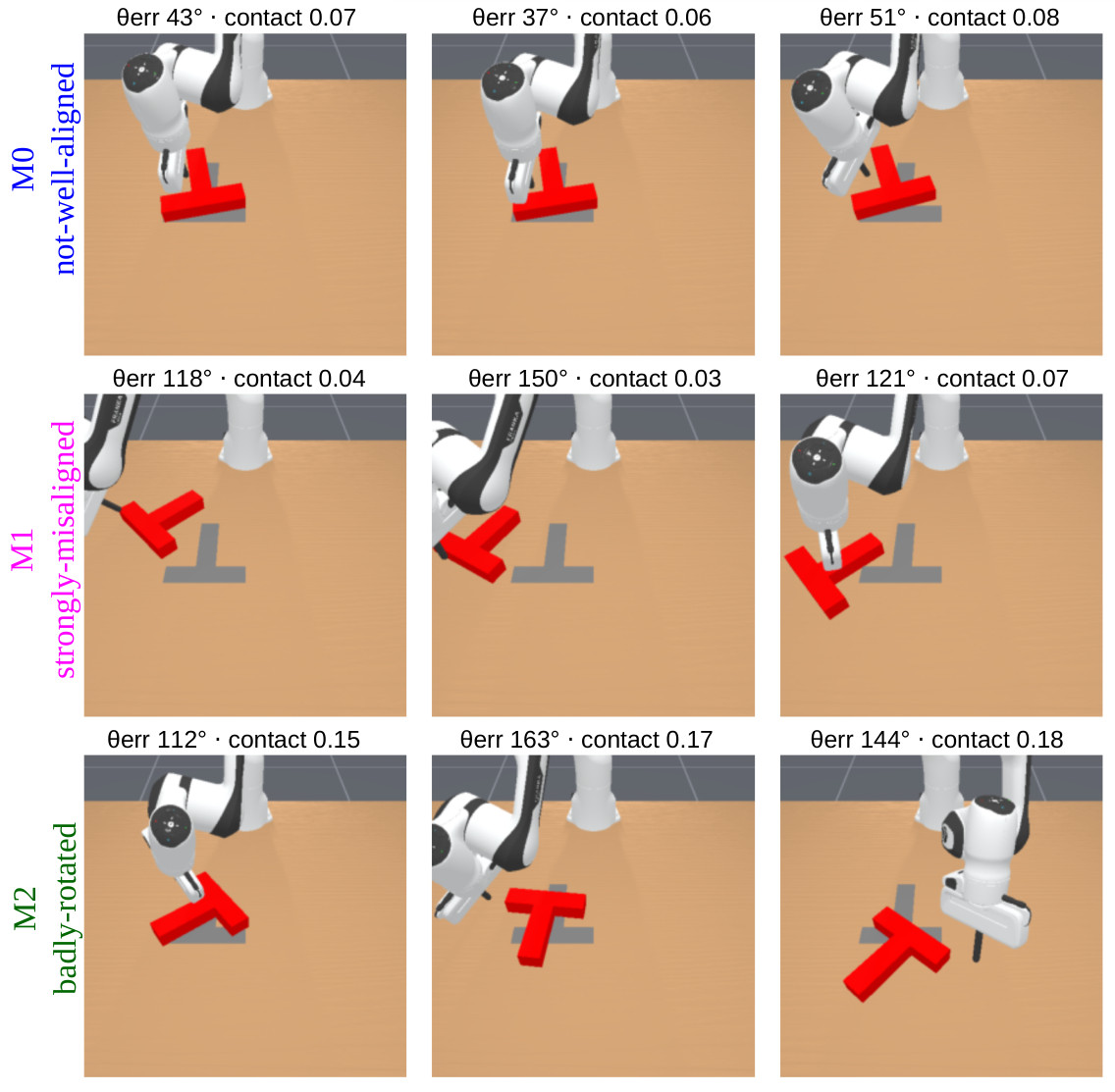}
\caption{Failure modes discovered on Push-T (image). Each row shows 3 members of a single geometric cluster. Labels are assigned by the naming pipeline
from the \mbox{task-store vocabulary}.}
\label{fig:failure_modes}
\end{figure}

\paragraph{Qualitative failure modes.}
Figure~\ref{fig:failure_modes} qualitatively shows that failures grouped by
geometry also exhibit similar configurations and semantic causes. These
clusters are the units over which DISEIL allocates demonstrations rather than
individual trajectories. Because the labels are produced by the naming pipeline
rather than independent annotation, the figure supports interpretability rather
than ground-truth cluster correctness.

\paragraph{Component knockouts.} Replacing mode-based allocation with greedy selection of the
highest-loss failure is the most damaging single change, at a mean of 4.37 points, while per-
demonstration goodness moves by only $+0.02$; the supporting components cost between 0.73 and 2.37
points each. With 1 arm per setting on 3 settings, this is a first indication that motivates
refining the allocation stage and running a broader ablation, and the supplement reports that
programme in full.

% !TeX root = ../main.tex
\section{What the study does and does not show}
\label{sec:scope}

This study is a controlled comparison carried out entirely in simulation. It supports the claim
that, holding the query rule and everything downstream of it fixed, choosing the failure mode and
the starting configuration raises held-out success rate on the 10 settings evaluated. It does not support claims beyond that,
and the evidence is limited in 4 respects.

\emph{The geometric descriptor is designed by hand and reads privileged state.} It is computed
only while a demonstration is being chosen and never reaches the policy that is deployed, but it
depends on object pose and end-effector geometry that a simulator supplies exactly. On a physical
robot the same quantities would have to be estimated from proprioception together with calibrated
RGB-D or multi-view cameras, object-pose estimation and task-specific perception. The size of the
resulting estimation error, and how much of the partition would survive it, is not established by
this evaluation. The descriptor also separates failures only when their cause is expressed in
geometry, and the cluster purity reported in the supplement is measured against the reasoning
model's own labels rather than against independent human annotation.

\emph{Most of the experts are not people.} GridWorld uses a human expert. Push-T uses a learned
PPO policy, and the RoboSuite tasks use scripted or motion-planned oracles. These produce clean,
low-variance demonstrations, so they do not exercise the noise, fatigue and varying skill of human
teaching \citep{belkhale2023dataquality,brown2019trex}. The budget also counts demonstrations
rather than the number of steps the expert controls. A prescribed demonstration is anchored at the
step where the policy first becomes unreliable, which lies earlier in the episode than the point
at which a DAgger gate hands over control, so the expert may have to drive the robot for longer.
How much extra expert time that costs is unexplored here, and comparing methods under a budget of
expert-controlled steps rather than of demonstrations would be a worthwhile direction to pursue.

\emph{The reasoning stage has a measurable cost.} Across the 5 settings that were instrumented,
it adds between 63 and 1{,}232 seconds and up to about 11{,}000 tokens per acquisition round. That
cost is paid once per round while a demonstration is being chosen, and never during policy
execution. For comparison, the work that DISEIL and the baseline both perform in the same round,
retraining the policy from scratch on the aggregated dataset and evaluating it on the held-out
set, takes between 783 and 1{,}491 seconds on the 3 RoboSuite settings. On those settings the reasoning
stage is therefore the smaller of the 2 costs. Whether it is worth paying depends on
what a demonstration costs, which is close to nothing for a scripted oracle and a person's time
otherwise.

\emph{The selector does not represent what the demonstration set already covers.} The cluster
memory records where corrections were requested in earlier rounds, not which behaviors the
collected demonstrations contain. DISEIL lacks any basis for separating a genuine gap in
supervision from repeated failure on behavior that has already been demonstrated and not learned.
The partitioning stage also becomes less active late in the budget, because a policy that has
improved leaves too few failures per round to cluster.

\section{Conclusion}

Under a fixed demonstration budget, sample efficiency depends on more than the rule that decides
when to call the expert. A further 2 decisions matter: which recurring failure mode receives the
next demonstration, and which configuration that demonstration starts from. DISEIL makes both explicitly, by grouping failures with a geometric descriptor, selecting a mode
on prevalence and severity, and letting the models write a request that the task constraints then
have to accept. It obtained the highest mean held-out success rate in all 10 settings evaluated, with a tie in 1, and its margin was largest at the smallest budget tested, 9.07 points at $B=10$
against 2.87 points at $B=20$. The component knockouts place most of the effect in the allocation stage rather
than in the models that describe the failures. If that holds under a broader ablation, the
underlying principle, spending a fixed budget across distinct recurring failure modes, should not
depend on the particular descriptor, clustering algorithm or language models used here.

We identify 3 limitations that point to directions for future work. First, every result reported here comes
from simulation. Closing the sim-to-real gap, by estimating the descriptor on a physical UR5 from
proprioception and calibrated cameras and measuring what the partition retains, is a prerequisite
for any claim about real robots. Second, reasoning over a single round of failures is a local view; a record of what the collected demonstrations already
cover would let the selector separate a behavior that is missing from one that has been
demonstrated and not learned. Third, when the teacher is a person rather than an oracle, a demonstration is no longer a single unit of the budget but an amount of someone's time, which makes
acquisition a cost-aware problem. In each case the open problem is the same. What supervision to collect is a design decision in its
own right, not a by-product of deciding when to interrupt.

\FloatBarrier
\bibliographystyle{plainnat}
\bibliography{references}

%%%%%%%%%%%%%%%%%%%%%%%%%%%%%%%%%%%%%%%%%%%%%%%%%%%%%%%%%%%%%%%%%%%%%%%%%%%%%%%
% Technical appendix, placed after the references.
%%%%%%%%%%%%%%%%%%%%%%%%%%%%%%%%%%%%%%%%%%%%%%%%%%%%%%%%%%%%%%%%%%%%%%%%%%%%%%%
\clearpage
\appendix

% Keep each appendix section's floats inside that section.
\makeatletter
\let\DISEIL@section\section
\renewcommand{\section}{\FloatBarrier\DISEIL@section}
\makeatother

% !TeX root = ../main.tex
% Technical appendix: the supplementary material, placed after the references.

\section{Implementation and Hyperparameters}

The implementation runs the 4 stages described in Section~\ref{sec:method} as a loop around an otherwise standard interactive imitation-learning harness.
Algorithm~\ref{alg:diseil} writes that loop out in full, at 1 demonstration per round.
Equation numbers inside it refer to Section~\ref{sec:method}.
Every run stores its exact prompts and replies, so the material quoted in
Appendix~K is a record rather than a reconstruction.

\begin{algorithm}[t]
\footnotesize
\caption{DISEIL with demonstration budget $B$}
\label{alg:diseil}
\begin{algorithmic}[1]
\Require Initial demonstrations $\mathcal{D}_0$; budget $B$; expert policy $\pi^\star$; retraining cadence $m$; held-out pool size $n$; loss threshold $\eta$; persistence window $K$; task knowledge base $\mathcal{K}$; context limit $\kappa$; proposal limit $J_{\max}$
\Ensure Trained policy $f_\theta$; $\mathcal{D}\gets\mathcal{D}_0$, $\mathcal{M}\gets\emptyset$, $r\gets0$
% Perceive
\State Fit $f_\theta$ on $\mathcal{D}$ by minimizing $\mathcal{L}_{\mathrm{BC}}$
\While{$|\mathcal{D}|-|\mathcal{D}_0|<B$}
    \State $r\gets r+1$: sample $n$ resets, roll out $f_\theta$, construct $\mathcal{F}_r$ (Eq.~2)
    \State $N \gets |\mathcal{F}_r|$
    \If{$\mathcal{F}_r=\emptyset$}
        \State \textbf{break}
    \EndIf
    \For{each $(\tau_i,\ell^{(i)}_{1:T_i})\in\mathcal{F}_r$}
        \State Trigger step $t_i^\star$ (Eq.~4)
        \State Anchor state $x_i^\star$; descriptor $\phi_i$ (Eq.~5)
        \State VLM/PM annotate $\tau_i$: root cause and phase via $\mathcal{K}$
    \EndFor
    % Partition
    \If{$N\geq4$}
        \State Standardize, select $k^\star$ by silhouette, cluster via $\mathcal{C}$ (Eq.~6)
    \Else
        \State 1 singleton mode per failure
    \EndIf
    \For{each mode $C$}
        \State Centroid $\bar{\phi}_C$; severity $\bar{L}_C$ (Eq.~7)
        \State Representative $\mathrm{rep}(C)$; memory-adjusted severity $\bar{L}^{\mathcal{M}}_C$
    \EndFor
    % Prioritize
    \State $C^\star\gets$ largest mode; select target $C_{\mathrm{tgt}}$ (Eq.~8)
    \State $S\gets\mathrm{rep}(C_{\mathrm{tgt}})$, highest-loss failure, fill to $\kappa$ (farthest-point)
    % Prescribe
    \State $\xi\gets\emptyset$, $a^{(0)}\gets\emptyset$
    \For{$j=1,\ldots,J_{\max}$}
        \State Generate proposal $\xi^{(j)}$ via PM, $g$ (Eq.~9)
        \If{$A_\theta(\xi^{(j)})$}
            \State $\xi\gets\xi^{(j)}$; \textbf{break}
        \Else
            \State $a^{(j)}\gets$ rejection reason
        \EndIf
    \EndFor
    \If{$\xi=\emptyset$}
        \State $\xi\gets$ nearest untried failure in $C_{\mathrm{tgt}}$
    \Else
        \State $\xi\gets\mathrm{rep}(C_{\mathrm{tgt}})$
    \EndIf
    \State Execute $\pi^\star$ from $\xi$; $\mathcal{D}\gets\mathcal{D}\cup\{d_r\}$ (Eq.~3)
    \State $\mathcal{M}\gets\mathcal{M}\cup\{(\bar{\phi}_{C_{\mathrm{tgt}}},\bar{L}_{C_{\mathrm{tgt}}})\}$
    \If{$r\bmod m=0$}
        \State Refit $f_\theta$ (Eq.~3)
        \State Recalibrate $\eta$
    \EndIf
\EndWhile
\State \Return $f_\theta$
\end{algorithmic}
\end{algorithm}

Table~\ref{tab:hyper} lists every value used in the reported runs. We highlight 3 choices here. Study A12 in Appendix~\ref{app:design} supports a context-set cap of 3 for the
Qwen3-32B instantiation used in the main experiments: 2 citations underperform,
whereas 5 provide no measurable improvement with this model. The model-sensitivity
comparison in A12b shows that the effect of additional context is not model-independent.
The re-prescription limit of 5 bounds the feasibility loop, and study A6 reports the
rate at which this limit triggers the deterministic fallback. The threshold $\eta$ is a
quantile of the policy's own training-loss distribution, recalibrated after each
retraining step so that the flagging rule adapts as the policy improves.

\begin{table}[t]
\centering
\caption{Every hyperparameter of the reported runs. The cluster-memory weight,
recency discount and bandwidth are held fixed within each task, and the memory is
enabled in every run reported in Section~\ref{sec:study}; study A1 in Appendix~\ref{app:knockouts} quantifies its
contribution.}
\label{tab:hyper}
\begin{tabular}{l>{\raggedright\arraybackslash}p{3.5cm}}
\toprule
Quantity & Value \\
\midrule
Budget $B$ & 20 demonstrations \\
Demonstrations per round $D$ & 1 \\
Rounds per run & $B/D = 20$ \\
Seeds, GridWorld & 9 \\
Seeds, robot tasks & 5 \\
Initial demonstrations $|\mathcal{D}_0|$ & 20 GridWorld, 20 Push-T, 12 Wipe, 8 Lift, 4 Door \\
Retraining & every demonstration, from scratch ($m = 1$) \\
\midrule
Descriptor width & 6 (study A10) \\
standardization & zero mean, unit variance \\
Partition & agglomerative, Ward linkage \\
Cluster count $k$ & $\arg\max$ mean silhouette \\
Sweep range & $[2,\,k_{\max}]$, with $k_{\max} = \max(2,\min(6,N-1))$ \\
Minimum failures to cluster & 4 \\
Context-set cap $\kappa$ & 3 (study A12; model sensitivity in A12b) \\
Diversity fill & farthest-point sampling \\
Re-prescription limit $J_{\max}$ & 5 \\
Fallback rule & nearest untried failure \\
\midrule
Threshold $\eta$ & quantile of the training-loss distribution, recalibrated at each retrain \\
Run length $K$ & consecutive steps above $\eta$ \\
\midrule
Vision-language model & Qwen3-VL \citep{bai2025qwen3vl} \\
Reasoning model & Qwen3 \citep{yang2025qwen3} \\
Prescription model & Qwen3 \citep{yang2025qwen3} \\
Frames per cited failure & 3 (start, $t^\star$, end) \\
Image-branch encoder & R3M \citep{nair2022r3m} \\
\bottomrule
\end{tabular}
\end{table}

\paragraph{Policies.} The framework requires only that the policy expose a per-step
loss, and it is instantiated with 3 policy classes to make that requirement
visible. The 4 robot tasks use diffusion policies
under both modalities \citep{chi2023diffusionpolicy}, with an R3M encoder supplying the
image branch \citep{nair2022r3m}. R3M supplies the policy's visual representation and
nothing else. It does not supply the clustering features, which are geometric in every
run, state and image alike.

\paragraph{Retraining cadence.} The policy is retrained from scratch on the aggregated
dataset after every acquired demonstration, on every task, GridWorld and robot alike,
so at $D = 1$ it is freshly fitted once per round, 20 times over the budget, and
$m = 1$ throughout. Retraining from scratch rather than warm-starting means each
reported gain is attributable to the acquired data and not to optimization carried over
between rounds, and each of the 20 rounds still draws a fresh rollout pool. The
cadence holds for every arm, so no arm is advantaged by the schedule. It also fixes the
goodness measurement of Section~\ref{sec:study}: because the scoring policy is retrained
after every demonstration, it is always the policy of the round, with no staleness on
any task.

\paragraph{Round accounting.} The seed counts are confirmed independently of the run
logs. Clustered rounds plus skipped rounds total 180 for each GridWorld setting and
100 for each robot setting, which is seeds times $B$ in both cases.

\paragraph{The starting-competence band.} The initial demonstration set is excluded
from the budget, and its size was not chosen freely. A policy's starting success rate
has to sit inside a band for the experiment to carry meaning. If the initial policy is
too weak, its rollouts fail everywhere, every configuration is a failure, the failure
set carries no structure for the descriptor to separate, and there is no allocation
problem to solve. If the initial policy is too strong, the failure set is empty or
nearly so, the budget has nothing to allocate, and every method converges to the same
place. The band between those 2 conditions is the regime in which a fixed budget can
be spent well or badly. Performance scales with the coverage of the
demonstrations a policy is trained on, while demonstrations added within a region
already covered saturate \citep{lin2024datascaling}, so the count is the lever that
places a task inside the band.

The principle is implemented as a behavior-cloning data-scaling sweep. A pool of
expert demonstrations is collected, behavior cloning is trained on each nested prefix
of the pool, each prefix is evaluated on the frozen held-out set, and the prefix whose
round-zero success rate is closest to a target near 50 percent is selected. That sweep
sets the $|\mathcal{D}_0|$ of Table~\ref{tab:hyper} and the resulting round-zero success rates span
43.1 to 67.2 percent, as recorded in Table~\ref{tab:main}. The count is set per task
rather than per modality, so the 2 modalities of a task do not begin at the same
success rate, and the image settings of Push-T and Door start slightly below the band,
at 43.3 and 43.1 percent. Lift begins the furthest above it, at 67.2 and 66.4, because
the smallest prefix that trains a stable policy on that task already clears the band.
Consequently, Lift operates near the performance ceiling in Table~\ref{tab:main}, which
limits the sensitivity of component-level comparisons on that task.

\paragraph{Language models and computing environment.}
The language-model components were served locally with vLLM using 3
NVIDIA H100 GPUs 80GB, with Qwen3-VL-32B for visual analysis and Qwen3-32B for
reasoning and prescription, both under 4-bit quantization. Policy training and
evaluation used NVIDIA A100, H100, H200, V100/V100L, or L40S GPUs with
Python~3.9, PyTorch~2.4.1, and CUDA~12.1.

\section{Tasks, Experts and the Geometric Descriptor}

\paragraph{Tasks.} A setting is a single task under a single observation modality, and the
evaluation covers 5 tasks under 2 modalities. The word \emph{mode} is used only
for a failure mode, which is a cluster of failures the framework discovers; an
observation modality is never called a mode.

GridWorld 5x5 is a discrete navigation task on a 5-by-5 grid with 3 obstacle
cells, in which an agent must reach a goal cell from a start cell. A* search and
breadth-first search enter the task only as the feasibility and path-validity checker
that decides whether a prescribed grid configuration admits an obstacle-free route from
start to goal \citep{hart1968astar, cormen2022algorithms}. Push-T is a planar pushing task in
which a manipulator must push a T-shaped block into a fixed goal pose. The task
originates in the implicit-behavior-cloning work that introduced it
\citep{florence2021implicitbc} and was popularized by the diffusion policy
\citep{chi2023diffusionpolicy}; the implementation used here is ManiSkill3's PushT-v1
\citep{tao2024maniskill3}. ManiSkill3 is the third release of a benchmark line whose
earlier releases carry a different task suite and do not contain Push-T
\citep{mu2021maniskill, gu2023maniskill2}, so the environment and the task are cited
separately. Lift, Wipe and Door are RoboSuite manipulation tasks on a UR5/UR5e arm
\citep{zhu2020robosuite}: lifting a cube from a table, wiping a randomised trail of
dirt markers from a surface, and pulling a door open past a hinge threshold.

\paragraph{Experts.} The expert differs by task, and each is named because the claim
that the demonstrations are correct by construction depends on it. On GridWorld the
expert is a human. On Lift, Wipe and Door the expert is a scripted oracle: an
open-loop motion-planner routine on Lift, a closed-loop routine reading the hinge
angle on Door, and a scripted wiping routine over the sampled marker path on Wipe. On
Push-T the expert is a policy trained by proximal policy optimization
\citep{schulman2017ppo}, used without modification, so the Push-T expert is learned
rather than scripted. It is an expert in the sense the framework requires, namely a
demonstrator whose trajectories are the target the policy is fitted to, and it is not
uniformly competent: the trained policy pushes in a single rotational direction only, so
configurations requiring the opposite rotation lie outside what it can demonstrate.
Those configurations are excluded by the workspace constraints stored in the task's
constraint store, which is the store described in Appendix~K.2.

\paragraph{The descriptor per task.} The descriptor is the same 6-dimensional vector
in both modalities of a task. Table~\ref{tab:descriptor} gives its components.
Position and orientation are the object's, not the end-effector's, on every task where
an object exists, because the configuration that determines whether a failure recurs is
the configuration of the thing being manipulated. Yaw enters through its sine and
cosine so the wrap at $\pm\pi$ does not create a false distance between 2 nearly
identical orientations. The 2 columns of a single task under the 2 modalities differ
because an image policy fails in different places, not because the features differ.

\begin{table}[t]
\centering
\caption{The 6 components of the geometric descriptor $\phi$ per task. The
descriptor is computed from privileged robot and object state at the flagged step and is
identical under both observation modalities. No output of any foundation model enters
it.}
\label{tab:descriptor}
\begin{tabular}{l>{\raggedright\arraybackslash}p{4.7cm}}
\toprule
Task & Components of $\phi$ \\
\midrule
GridWorld & agent cell (2), signed offset to goal (2), progress, Manhattan distance to goal \\
Push-T & block planar position (2), $\sin\psi$, $\cos\psi$, progress, end-effector-to-block distance \\
Lift & cube planar position (2), progress, gripper-to-cube distance, gripper height, grasp indicator \\
Door & door-frame position (2), frame yaw, normalized hinge angle, end-effector-to-handle distance, progress \\
Wipe & remaining-dirt centroid (2), proportion wiped, end-effector-to-centroid distance, markers remaining, progress \\
\bottomrule
\end{tabular}
\end{table}

\section{Baselines and the Uniform-Random Control}

The evaluation runs 6 comparison methods. Of these, 5 are published interactive
imitation-learning methods and share a single skeleton: roll out the current policy, read a
scalar signal, hand control to the expert when the signal crosses a threshold,
aggregate the expert's labels and retrain \citep{ross2011dagger}. They differ only in
the signal. SafeDAgger learns a classifier that predicts when the policy is about to
deviate from the expert \citep{zhang2017safedagger}. DropoutDAgger reads the spread of
a Monte-Carlo dropout ensemble of the learner's action distribution
\citep{gal2016dropout, menda2017dropoutdagger}. EnsembleDAgger reads the variance of an
explicit ensemble combined with an action-discrepancy term
\citep{lakshminarayanan2017ensembles, menda2019ensembledagger}. ThriftyDAgger combines
a novelty estimate with a learned risk estimate under a target switching rate
\citep{hoque2021thriftydagger}. Diff-DAgger uses a diffusion policy's own per-step
training loss as the uncertainty signal \citep{lee2025diffdagger} and is run on the
robot tasks only, where the policy is a diffusion policy. Those 5 are the DAgger
family and are labeled as such in every comparison. Diff-DAgger's use of the per-step
diffusion loss as an uncertainty signal is its own contribution; DISEIL uses that
signal for failure localization and also compares against it as a baseline, and both
facts are stated wherever the signal appears.

STAGGER \citep{li2025interactivehybridil} is the sixth comparison method and is reported
only on GridWorld in Table~\ref{tab:main}. Its annotation model queries expert actions
at individual states rather than acquiring full trajectory demonstrations, which makes it
incompatible with the fixed demonstration budget on the robotic manipulation tasks. A separate uniform-random control, implemented in this
project, corrects 1 uniformly chosen recorded failure each round, with no gate, no
descriptor and no allocation. That control is study A2 in Appendix~F, where it answers the
question of whether unstructured failure replay alone explains DISEIL's gains.

\section{Extended Derivations and Rationale}

Each rule of the framework is stated in Section~\ref{sec:method}. The
remainder of this section provides the design rationale for those choices.

\paragraph{Why the first crossing and not the peak.} The flagged step is the first step
at which the per-step loss exceeds $\eta$ for $K$ consecutive steps, and the peak is
used only when no crossing occurs. Relative to a peak-loss anchor, the first sustained
crossing identifies the onset of unreliability. It also anchors the descriptor closer to the
configuration in which the failure begins, rather than to the later state produced
after errors have compounded.

\paragraph{The bound on the cluster sweep, and when it is skipped.} The cluster count is chosen
by maximum mean silhouette over $k \in [2, k_{\max}]$ with
$k_{\max} = \max(2, \min(6, N-1))$. The lower bound of 2 is forced because a
silhouette is undefined at $k = 1$. The cap at 6 is not a claim that 6 modes exist;
it is a consequence of the sample size, since the failure sets are small and shrink over
the budget. The cap avoids evaluating excessively fine partitions of small failure
sets. When fewer than 4 failures remain, the sweep is skipped
entirely, each failure becomes its own singleton, and the round is allocated by the
fallback rule. Study A14 in Appendix~H reports how often that happens and study A16
reports when in the budget it happens.

\paragraph{The size constraint on the target.} The target is the mode of highest
mean peak loss among those whose size is within 1 member of the dominant mode's. The
constraint exists because mean peak loss is a mean over a set that can be very small.
Without it, a single badly failed episode that lands far from every other failure
becomes its own cluster, carries the highest mean peak loss by construction, and
captures the round's demonstration. The constraint keeps the target inside the bulk of
the round's failures, so the framework spends the round on a mode the policy is
actually producing rather than on an outlier. It is the reason the prioritization rule
is not simply the maximum of $\bar{L}_C$.

\paragraph{Why feasibility is checked before the expert is called.} A prescription is a
request for a configuration of the world, and a language model asked for a
configuration will sometimes ask for one the world cannot produce: an object outside
the reachable set, a pose outside the spawn range, a grid layout with no path from
start to goal. The store of environmental constraints is what makes such a request
checkable. It is not a document store to be retrieved from in the manner of
retrieval-augmented generation \citep{lewis2020rag, edge2024graphrag}; it is closer to
the explicit, queryable environment and action knowledge of a robot knowledge base
\citep{tenorth2013knowrob}. A violation is returned to the model as feedback and a
revised prescription is demanded until a feasible one is produced or the limit
$J_{\max}$ is reached. A failed attempt consumes no budget, because the budget counts
demonstrations collected and not prescriptions proposed. Propose-verify-revise against
an external checker is a standard pattern \citep{liu2023llmp, chen2024autotamp}; what
is new is not the checker but the object being verified, which is a request for a
training demonstration that has not been collected rather than a plan to be executed.

\paragraph{The division of labor between the 2 model roles.} The descriptor is
computed from robot and object state and the frames go to a vision-language model, and the 2
never mix. Vision-language models are competent at naming a cause when they are given
structured evidence \citep{duan2025aha, liu2023reflect} and unreliable at metric and
spatial reasoning from pixels alone \citep{chen2024spatialvlm, fu2024blink}. The
framework therefore asks them for the cause and computes the geometry itself. A mode's
name is the majority root cause among its members, taken from a closed taxonomy stored
in the task's constraint store rather than invented by the model, because a partition
returns integers and a method that reports a failure in mode 2 has said nothing.

\paragraph{The solvability screen.} The second screen asks whether the current policy
already solves the prescribed configuration and revises it if so, on the reasoning that
a demonstration of something the policy can already do teaches nothing. Its nearest
relatives are reverse curriculum generation \citep{florensa2017reversecurriculum} and
reset learning \citep{eysenbach2018leavenotrace}. It is a design element only.

\paragraph{The derived quantity.} The term,
\emph{Margin retained} is the fraction of DISEIL's advantage over the strongest
baseline that survives an ablation,
$(\text{ablated} - \text{best baseline}) / (\text{full} - \text{best baseline})$,
expressed as a percentage. It is reported alongside the raw damage because a component
whose removal costs 2 points where the margin is 3 points is a different object
from a component whose removal costs 2 points where the margin is 10. A value near
100 percent indicates that most of the margin remains, a value near zero indicates that
little of the margin remains, and a negative value indicates that the ablated system
falls below the strongest baseline.

\section{The Ablation program: Scope and Conventions}

The program comprises 17 studies. A1 to A12 remove or vary 1 component at a
time. A13 to A17 are diagnostics: they measure a property of the running system rather
than knock a component out of it. A12b is an additional model-sensitivity comparison
that tests whether the context-set-size result transfers across prescription models.

The studies are run and reported on 3 settings, chosen to span the 3 policy
classes and both observation modalities: GridWorld (image), where the policy is a
convolutional network; Push-T (state), where it is a state diffusion policy; and Door
(image), where it is an image diffusion policy. Every per-setting number below is 1
of those 3, quoted in that order, and every aggregate is the mean over the 3
and is labeled as such.

The unit of analysis is the setting. Each study is reported as its 3 per-setting
values, the sign they share and the mean over the 3.

\section{Knockouts, A1 to A8}
\label{app:knockouts}

\begin{figure}[t]
\centering
\includegraphics[width=0.74\linewidth]{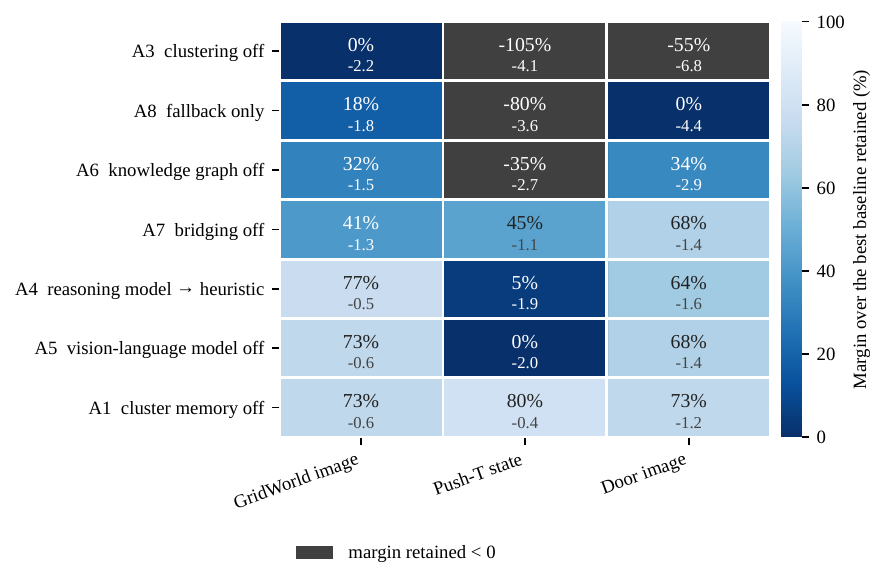}
\caption{Cost of each knockout in points of final success rate against full DISEIL,
averaged over the 3 ablation settings. The partition knockout is the most damaging;
the plot shows the relative contribution of the remaining components.}
\label{fig:knockoutsummary}
\end{figure}

\begin{figure}[t]
\centering
\includegraphics[width=0.74\linewidth]{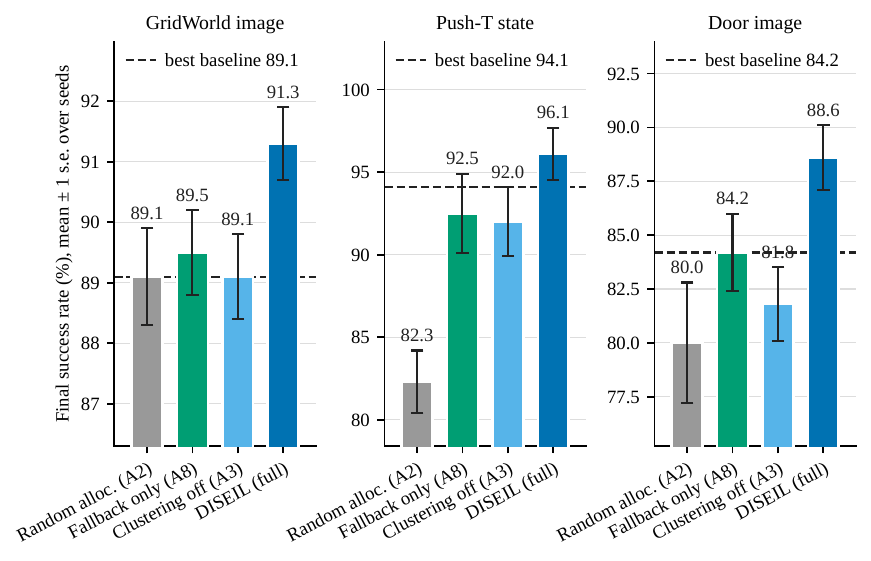}
\caption{The allocation ladder on the 3 ablation settings. Bars are the mean final
success rate over seeds with 1 standard error, for uniform-random allocation over
recorded failures (A2), the deterministic nearest-untried fallback rule promoted to the
whole method (A8), clustering removed in favor of greedy worst-loss selection (A3),
and full DISEIL. The dashed line is the strongest baseline in that setting. Budget
$B = 20$, 1 demonstration per round.}
\label{fig:ladder}
\end{figure}

The knockouts were run from the bottom of the system upwards and are reported in that
order: first the 2 controls that establish what unstructured and unreasoning
allocation can already do, then the components the method claims for itself.
Table~\ref{tab:knockouts} is the full grid and Table~\ref{tab:ladder} is the ladder in
absolute terms. Figure~\ref{fig:ladder} plots the ladder and
Figure~\ref{fig:knockoutsummary} summarizes the per-knockout cost.

\begin{table}[t]
\centering
\caption{Every knockout, on the 3 ablation settings, ordered by mean damage.
Values are the change in final success rate in points against full DISEIL. Mean margin
retained is the mean over the 3 settings of
$(\text{ablated} - \text{best baseline}) / (\text{full} - \text{best baseline})$, as a
percentage. A negative value means the ablated system finished, on average, beneath the
baseline it was built to beat. GW is GridWorld (image), PT is Push-T (state) and Dr is
Door (image).}
\label{tab:knockouts}
\footnotesize
\setlength{\tabcolsep}{4pt}
\begin{tabular}{llrrrrr}
\toprule
 & & \multicolumn{3}{c}{$\Delta$ success rate (points)} & & Mean margin \\
\cmidrule(lr){3-5}
Study & Component knocked out & GW & PT & Dr & Mean & retained (\%) \\
\midrule
A3 & the clustering step & $-2.2$ & $-4.1$ & $-6.8$ & $-4.37$ & $-53.2$ \\
A8 & the prescription, fallback rule promoted & $-1.8$ & $-3.6$ & $-4.4$ & $-3.27$ & $-20.6$ \\
A6 & the environmental constraints & $-1.5$ & $-2.7$ & $-2.9$ & $-2.37$ & $10.3$ \\
A4 & the prescription model & $-0.5$ & $-1.9$ & $-1.6$ & $-1.33$ & $48.6$ \\
A5 & the vision-language model & $-0.6$ & $-2.0$ & $-1.4$ & $-1.33$ & $47.0$ \\
A7 & bridging placement & $-1.3$ & $-1.1$ & $-1.4$ & $-1.27$ & $51.4$ \\
A1 & the cluster memory & $-0.6$ & $-0.4$ & $-1.2$ & $-0.73$ & $75.1$ \\
\bottomrule
\end{tabular}
\end{table}

\begin{table}[t]
\centering
\caption{The allocation ladder in absolute terms: final held-out success rate (per
cent) on the 3 ablation settings. A2 replaces allocation with a uniform draw over
recorded failures, A8 promotes the deterministic fallback rule to the whole method,
and A3 keeps the loss signal and removes the mode structure. GW is GridWorld (image),
PT is Push-T (state) and Dr is Door (image).}
\label{tab:ladder}
\begin{tabular}{llrrr}
\toprule
 & Arm & GW & PT & Dr \\
\midrule
A2 & uniform-random allocation & 89.1 & 82.3 & 80.0 \\
A8 & fallback rule only & 89.5 & 92.5 & 84.2 \\
A3 & greedy worst-loss & 89.1 & 92.0 & 81.8 \\
 & full DISEIL & \textbf{91.3} & \textbf{96.1} & \textbf{88.6} \\
\bottomrule
\end{tabular}
\end{table}

\paragraph{A2, uniform-random allocation.}
This control selects 1 recorded failure uniformly at random in each round and requests an expert correction for that failure. It uses no geometric descriptor, failure-mode partition, cluster memory, or prescription model. It is distinct from the STAGGER baseline reported in the main results and is evaluated on the 3 ablation settings. Uniform-random allocation achieves final success rates of 89.1, 82.3, and 80.0 on GridWorld (image), Push-T (state), and Door (image), respectively, compared with 91.3, 96.1, and 88.6 for DISEIL. On the 2 robot settings, it also underperforms the strongest uncertainty-gated baseline. On GridWorld (image), it performs similarly to the gated baselines but remains below DISEIL.

Uniform sampling over individual failures implicitly selects each failure mode in proportion to its prevalence in the current failure set. Consequently, common modes receive a larger share of the fixed demonstration budget, whereas rare modes receive fewer opportunities for correction and may remain unaddressed. A2 therefore distinguishes the effect of structured failure-mode allocation from the benefit of failure replay alone. Its lower performance, particularly on Push-T (state) and Door (image), shows that revisiting recorded failures is insufficient to recover DISEIL's gains; the allocation of demonstrations across discovered failure modes provides the additional benefit.

\paragraph{A8, the deterministic fallback rule.}
If the prescription model fails to produce a feasible request within 5 attempts, DISEIL selects an untried recorded failure using the geometric descriptor and requests an expert correction from that failure. A8 applies this deterministic rule in every round, replacing the prescription model throughout the run. It achieves final success rates of 89.5, 92.5, and 84.2 on GridWorld (image), Push-T (state), and Door (image), respectively, corresponding to reductions of 1.8, 3.6, and 4.4 percentage points relative to full DISEIL. It retains 18.2\%, $-80.0$\%, and 0.0\% of DISEIL's margin over the strongest baseline. Thus, A8 remains above the strongest baseline on GridWorld, falls below it on Push-T, and matches it on Door.

The deterministic rule promotes geometric diversity by avoiding repeated correction of already addressed failures. However, it does not explicitly represent recurring failure modes, prioritize modes by severity, or maintain mode-level correction history. Its lower performance across all 3 settings shows that geometric failure selection provides a useful fallback but does not recover the benefit of DISEIL's complete failure-mode allocation and prescription pipeline.

\begin{figure}[t]
\centering
\includegraphics[width=0.74\linewidth]{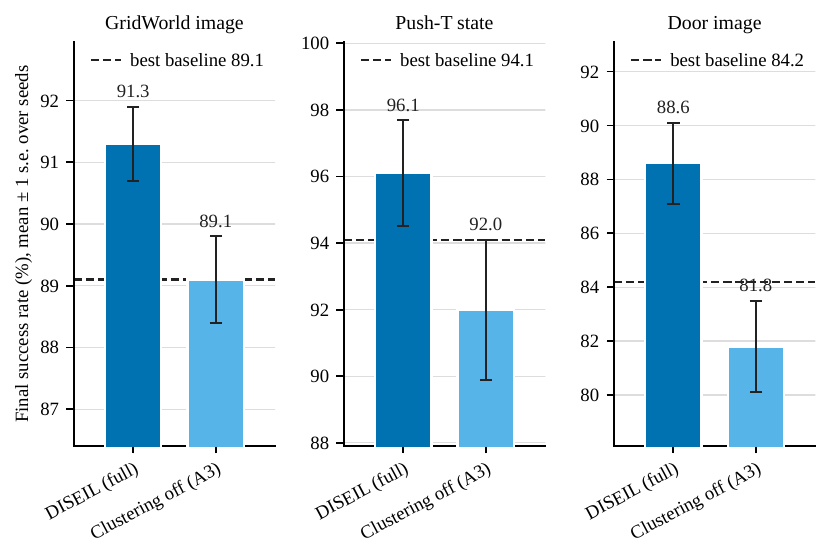}
\caption{Success rate with clustering removed. Final success rate for full DISEIL and
for greedy worst-loss selection (A3) on the 3 ablation settings, with the strongest
baseline as a dashed line.}
\label{fig:gainalloc}
\end{figure}

\paragraph{A3, the clustering step, and the dissociation.} The third rung removes the
clustering step and targets, each round, the single failure with the highest peak
per-step loss. The loss signal is kept and the failure-mode structure is removed.
Success falls by $2.2$, $4.1$ and $6.8$ points, a mean of $4.37$, and the margin retained
collapses to $0.0$, $-105.0$ and $-54.5$ percent, a mean of $-53.2$. On Push-T (state)
and Door (image) the ablated system falls beneath its own best baseline, $92.0$ against
$94.1$ and $81.8$ against $84.2$, and on GridWorld (image) it lands exactly on it. Removing
the clustering step erases the whole margin and on 2 of the 3 settings turns it
negative, the largest damage of any knockout in the program.
Figure~\ref{fig:gainalloc} plots the arm against its baselines.

A3 provides the central mechanistic evidence because per-demonstration goodness and final success respond differently to the same intervention. Removing clustering changes goodness by $(+0.02)$, $(+0.16)$, and $(-0.13)$ across the 3 ablation settings, yielding a mean change of $(+0.02)$. Thus, greedy worst-loss selection remains as effective as full DISEIL at acquiring demonstrations with high pre-retraining policy loss. This is expected because A3 explicitly selects the individual failure with the highest peak loss, whereas DISEIL may select a lower-loss failure to distribute supervision across distinct failure modes. Despite comparable goodness, A3 reduces final success rate by $4.37$ percentage points on average. This dissociation shows that the value of a fixed demonstration budget depends not only on the novelty of each demonstration in isolation, but also on how supervision is distributed across the policy's recurring failures.

\begin{figure}[t]
\centering
\includegraphics[width=\linewidth]{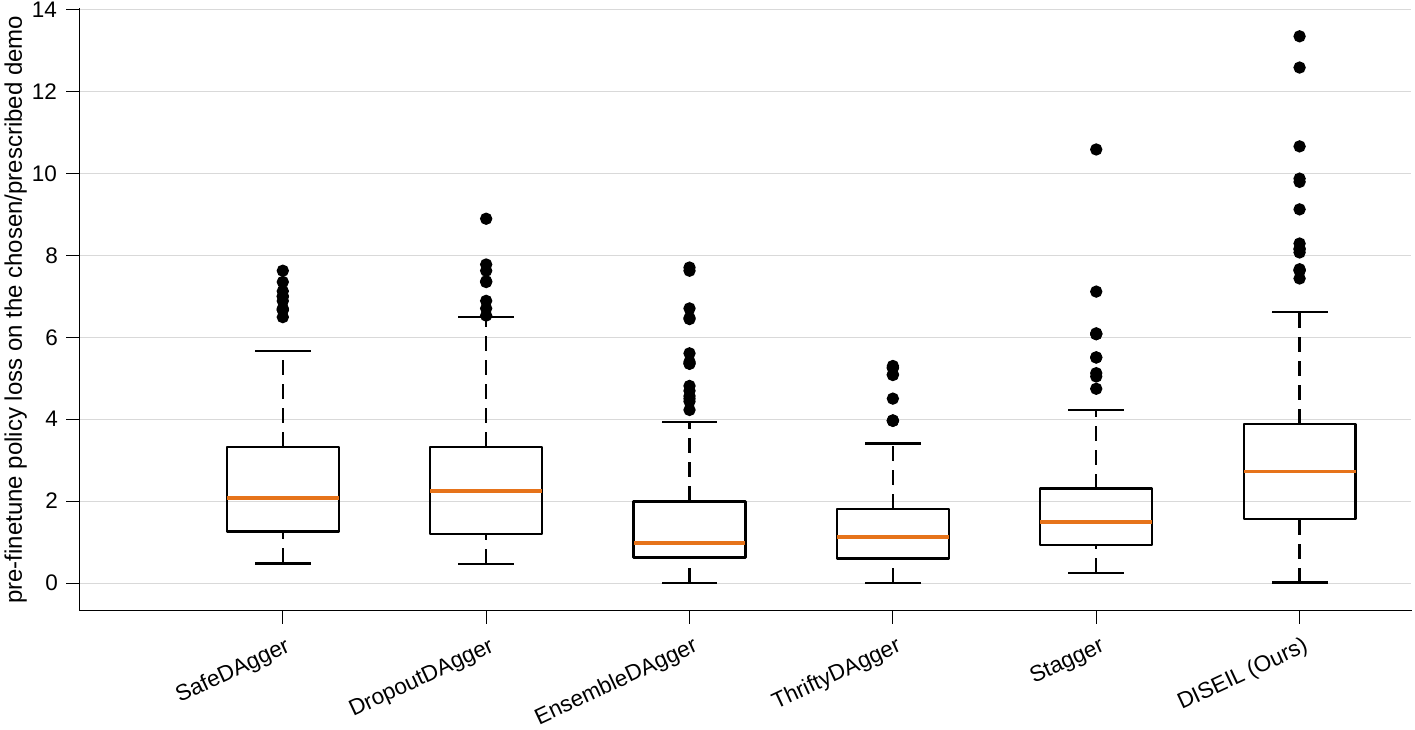}
\caption{Per-demonstration goodness, the policy's per-step loss on each newly
acquired demonstration measured before retraining on it. The distributional view of the
goodness table of Table~\ref{tab:infogain}. Diff-DAgger is run on the robot tasks only.}
\label{fig:infogain}
\end{figure}

Peak loss characterizes the severity of an individual trajectory; it does not characterize the coverage of the complete failure set. Greedy worst-loss selection can repeatedly prioritize high-loss failures from a limited region of the state space, leaving other recurring modes less frequently addressed. Under this interpretation, the selected demonstrations may remain individually informative while providing less diverse supervision collectively. DISEIL instead uses failure-mode clustering and memory to distribute demonstrations across distinct regions of the failure distribution. Figure~\ref{fig:infogain} presents the per-setting distribution of the goodness values summarized in Table~\ref{tab:infogain}.

Section~\ref{sec:study} reports Diff-DAgger as the primary comparison for goodness because Diff-DAgger selects expert interventions using the same per-step diffusion loss from which the metric is computed. It is therefore the baseline most directly aligned with acquiring high-loss demonstrations. The remaining query-gated baselines and the uniform-random control use different selection signals and are omitted from that table for concision. Across their applicable settings, DISEIL achieves higher mean goodness than the query-gated baselines and also outperforms the uniform-random control wherever it is evaluated.

A3 removes the descriptor and memory together with clustering because these components
operate on the discovered modes. Its result therefore identifies the contribution of
the integrated allocation \emph{stack}, rather than the clustering algorithm in
isolation.

\paragraph{A6, the constraints the prescription is checked against.} The prescription
model proposes a configuration; constraints are retrieved from the task's store, which
holds workspace bounds, reachability, object and spawn ranges and controller limits as
structured key-value knowledge; the proposal is checked against them; a violation is
returned to the model as feedback and a revised proposal is requested until a feasible
one is produced. A6 removes the store from both the vision-language and the reasoning
prompts, so the loop has nothing to verify against.

\begin{figure}[t]
\centering
\includegraphics[width=0.74\linewidth]{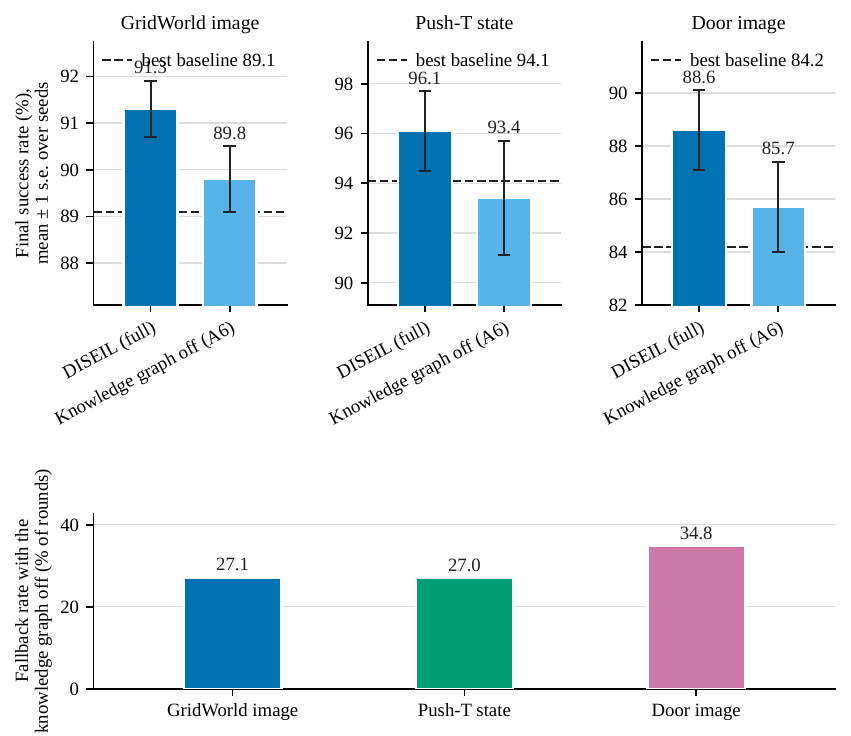}
\caption{Grounding and feasibility. Top row: final success rate for full DISEIL and
with the environmental constraints removed from the prompts (A6), against the strongest
baseline. Bars are means over seeds with 1 standard error, including the A6 bar.
Bottom row: the share of rounds that fall back to the deterministic rule when the
constraints are removed.}
\label{fig:grounding}
\end{figure}

The cost is 1.5, 2.7 and 2.9 points, a mean of 2.37, and 31.8, $-35.0$ and 34.1 per
cent of the margin is retained, a mean of 10.3. A6 is the third most damaging knockout
and it costs nearly twice what the prescription model itself is worth. The fallback
rate rises to 27.1, 27.0 and 34.8 percent of rounds, which means roughly 5 to 7
of the 20 rounds are spent on a fallback correction rather than a prescribed one, a
direct loss of a quarter to a third of the budget. Figure~\ref{fig:grounding} shows
both halves of the measurement.

The fallback rate and success-rate change need not scale one-for-one: a fallback round
still acquires the nearest untried failure, so its cost is the difference between a
prescribed and a fallback correction rather than the loss of the entire round. Together,
A6 and A8 show that the environmental store contributes by keeping prescriptions
feasible while the fallback preserves useful supervision when re-prescription is
exhausted. Because A6 removes the complete task store, its 2.37-point effect quantifies
the structured constraint set as a whole rather than any individual field.

\begin{figure}[t]
\centering
\includegraphics[width=0.74\linewidth]{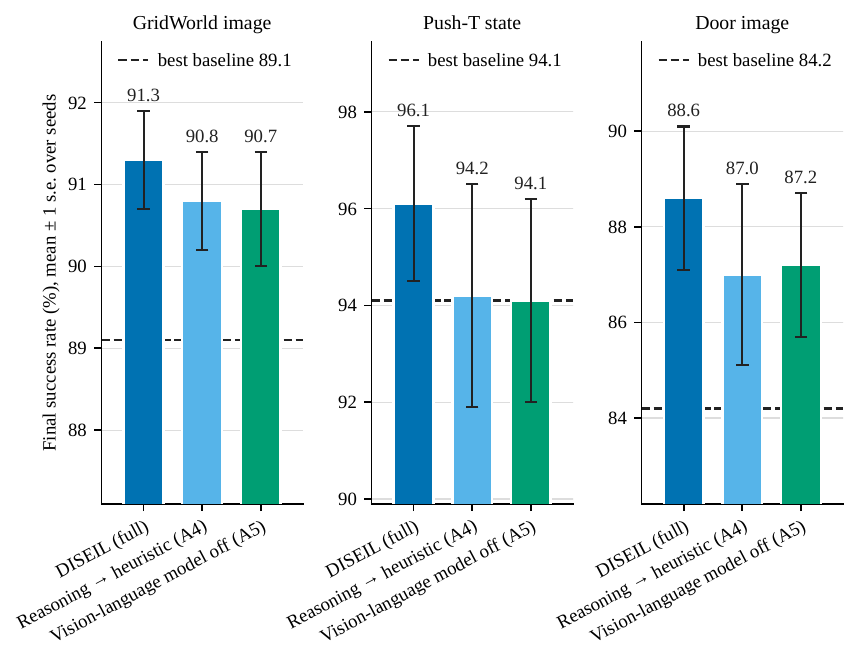}
\caption{The prescription model and the vision-language model. Final success rate for
full DISEIL, for the prescription model replaced by the dominant-representative
heuristic (A4), and for the vision-language model removed (A5), with 1 standard error
over seeds and the strongest baseline as a dashed line.}
\label{fig:reasoning}
\end{figure}

\paragraph{A4 and A5, the 2 model roles.} The \emph{prescription model} is the
text-only prescription model of Section~\ref{sec:method}, and it fills 2 roles. In the first it assigns a
root cause and a trajectory phase to each failure, once per failure, from the taxonomy
stored in the task's constraint store. In the second it turns the selected mode and its
context set into the round's demonstration request, once per round. A4
replaces the second of those with the deterministic rule ``always target the dominant
cluster representative'', operating on the same geometric clusters, so that the
comparison isolates the prescription decision and not the partition; the root-cause
labeling role is retained in A4 and still runs. A5 removes the vision-language model,
leaving the prescription model with the geometric descriptor and the root-cause taxonomy.
The 2 roles are removed together nowhere in the program, and the phrase
\emph{reasoning stack} names both roles of the prescription model plus the
vision-language call taken
together. Figure~\ref{fig:reasoning} plots both arms.

A4 costs 0.5, 1.9 and 1.6 points and A5 costs 0.6, 2.0 and 1.4, so both average 1.33
points and both retain roughly half the margin, 48.6 percent for A4 and 47.0 for A5.
Both components produce consistent improvements across the 3 settings. Their
similar effect sizes do not support a ranking between them; both are largest on Push-T
(state) and smallest on GridWorld (image).

The explanation is structural. Clustering is geometric in every run, state and image
alike, and it consumes no output from any foundation model. By the time either model is
called, the target region of the failure distribution has already been selected by the
descriptor and the memory. The
prescription model chooses the form of the correction inside a region that was selected
without it. The results support DISEIL's modular design: geometric allocation supplies
the primary selection mechanism, while the prescription and vision-language models
refine how the selected region is corrected. They also establish a reduced-compute
variant: retaining geometric clustering, memory, and the deterministic representative
heuristic continues to outperform every baseline in the evaluated settings.

A4 compares the prescription model with a strong allocation-aware heuristic: ``always
target the dominant cluster representative.'' This heuristic inherits the memory's
rotation, while the full model additionally supports bridging. A5 shows that visual
evidence remains beneficial even with state-based policies: its largest effect occurs on
Push-T (state), where a shared terminal geometry can arise from distinct processes such
as pushing on the wrong face, losing contact, or over-rotating. On Door, where terminal
geometry more directly identifies the failure cause, the measured visual contribution is
smaller.

\begin{figure}[t]
\centering
\includegraphics[width=0.74\linewidth]{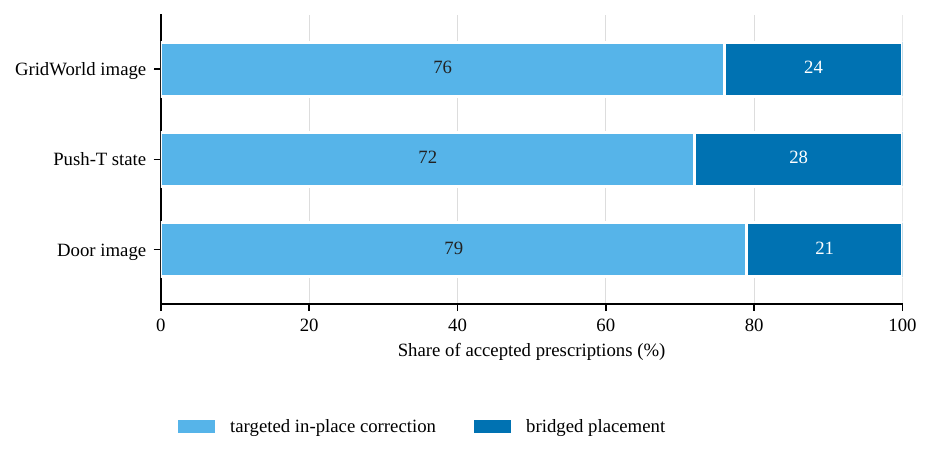}
\caption{Targeted and bridged prescriptions (study A7). Share of accepted
prescriptions on each of the 3 ablation settings that are a targeted in-place
correction and that are a bridged placement.}
\label{fig:bridging}
\end{figure}

\paragraph{A7, bridging placement.} Bridging is the only part of the prescription that
changes the environment configuration rather than selecting an episode, and it is the
mechanism by which a prescription can be made easier than the failure it addresses.
Disabling it costs 1.3, 1.1 and 1.4 points, a mean of 1.27, and retains 40.9, 45.0 and
68.2 percent of the margin. That is proportionate to how often the arm is used: the
bridged share of accepted prescriptions, shown in
Figure~\ref{fig:bridging}, is 24, 28 and 21 percent. A component used in a quarter of
rounds cannot produce a large aggregate effect unless the rounds in which it is used
are the decisive ones, and the 3 settings do not order the damage the way they
order the share: the largest damage is on the setting with the middle share. What
matters is therefore which rounds use bridging, rather than its aggregate frequency
alone.

\paragraph{A1, the cluster memory.} Switching the memory off, $\lambda = 0$, costs 0.6,
0.4 and 1.2 points, a mean of 0.73, and retains 72.7, 80.0 and 72.7 percent of the
margin, a mean of 75.1. The memory suppresses a cluster that has just received a
demonstration, so the following round is pushed onto a different one. A1 quantifies
this contribution after partitioning. The measured effect is task-dependent: 0.4 points on Push-T
(state) and 1.2 on Door (image). These consistent improvements quantify the memory's
complementary contribution after the partition has been formed. The candidate set of
near-dominant modes is a singleton in 56 to 84 percent of rounds on the settings with
sufficient telemetry; in the remaining rounds, the memory can redirect supervision
away from recently corrected regions.

\paragraph{The ordering.} Ranked by mean damage, the 7 knockouts are: clustering
($-4.37$ points, $-53.2$ percent of the margin retained), the fallback rule promoted to
the whole method ($-3.27$, $-20.6$), the constraint store ($-2.37$, $10.3$), the
prescription model and the vision-language model ($-1.33$ each, $48.6$ and $47.0$),
bridging placement ($-1.27$, $51.4$) and the cluster memory ($-0.73$, $75.1$). The
ordering shaped the rest of the program in 2 ways. The clustering step, which the
framework instantiates as a generic partition, produces the largest measured
contribution, establishing mode-based allocation as the principal driver. The cluster
memory provides a smaller, consistent, task-dependent complement to that partition.

\section{Design Choices, A9 to A12}
\label{app:design}

Given that allocation is the mechanism, the next family of studies asks whether
\emph{this} descriptor, \emph{this} cluster count and \emph{this} context set provide effective design choices for allocating the fixed
demonstration budget. The additional A12b comparison asks whether the context-set result
transfers across prescription models.

\paragraph{A10, the width of the descriptor.} The descriptor is designed, not learned,
and A10 validates its width using mean silhouette, a cluster-separation criterion
independent of success rate \citep{rousseeuw1987silhouette}. Features are removed from
and added to the descriptor, and the silhouette is measured over every clustering
round.

\begin{figure}[t]
\centering
\includegraphics[width=0.74\linewidth]{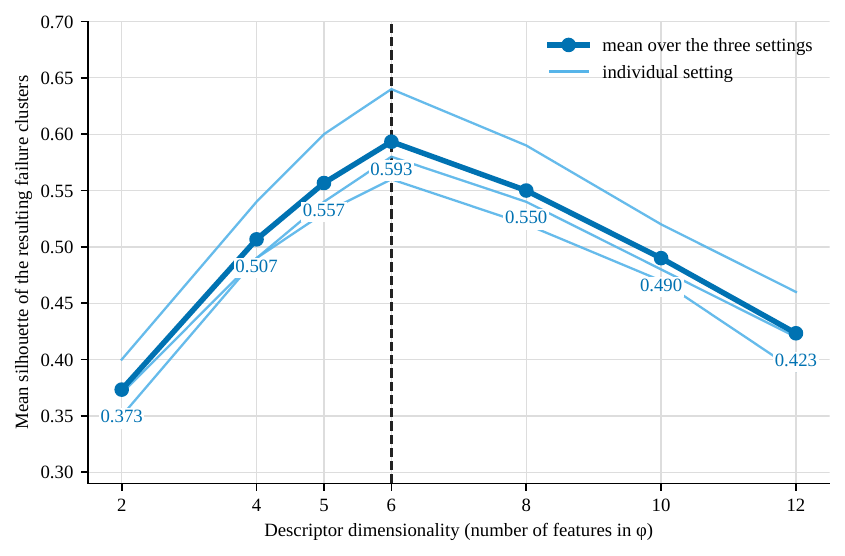}
\caption{Mean silhouette of the failure clusters against the dimensionality of the
geometric descriptor, with 1 line per ablation setting and the mean over the 3. The
dashed vertical line marks the 6-dimensional descriptor the framework uses.}
\label{fig:descdim}
\end{figure}

The curve in Figure~\ref{fig:descdim} is a clear inverted U with a single interior maximum, and the 6-dimensional
descriptor is the highest-scoring variant in each of the 3 settings. Below 6 dimensions the descriptor discards information that distinguishes
modes, and the largest single step in the whole sweep, $+0.133$ in mean silhouette, is
the step from 2 dimensions to 4, which adds orientation. Position alone cannot
separate a Push-T failure in which the block is in the right place at the wrong angle
from one in which it is in the right place at the right angle and the pusher lost
contact, so those failures collapse into a single cluster and the cluster's silhouette is
poor. Adding task progress, worth $+0.050$, and contact distance, worth $+0.037$, each
buy less, which is consistent with orientation being the dominant discriminator on
these tasks.

Above 6 dimensions, the observed decline is consistent with distance concentration:
as dimensionality rises, pairwise distances become less discriminative for clustering
small failure sets. The added variants include end-effector velocity at 8
dimensions and a joint-angle summary by 12 dimensions. Because the instrumented
setting contains 42 failures in round 1 and 2 by round 20, the compact
6-dimensional representation provides the strongest measured separation at the
sample sizes encountered by the method.

A10 also confirms an implementation detail. Clustering is geometric for every run, state
modality and image modality alike, and the descriptor is the same 6-dimensional
vector in both. Silhouette measures geometric separation; the complementary purity
diagnostic in Appendix~H evaluates agreement with the system's root-cause labels. Thus,
A10 establishes that 6 dimensions is the strongest variant within the evaluated
descriptor family, while A13 characterizes the semantic consistency of the resulting
clusters.

\paragraph{A9 and A12, the context set.} The context set given to the prescription model
contains 3 cited failure episodes chosen by 3 rules: the forced representative
of the target cluster, the worst-peak-loss seed, and a farthest-point-sampling fill for
diversity \citep{eldar1997fps}. Written out, the set $S$ referred to in Section~\ref{sec:method}'s
prioritize stage is built from the target mode $C_{\mathrm{tgt}}$ as
\begin{equation}
\label{eq:context}
\begin{aligned}
S_0 &= \big\{\mathrm{rep}(C_{\mathrm{tgt}})\big\} \cup \Big\{\arg\max_{i \in
C_{\mathrm{tgt}}} \mathrm{peak}_i\Big\}, \quad S \leftarrow S_0, \\
S &\leftarrow S \cup \Big\{ \arg\max_{i \in C_{\mathrm{tgt}} \setminus S} \\
\min_{j \in S} \ \big\| \tilde{X}_i - \tilde{X}_j \big\|_2 \Big\} \\
&\qquad \text{ until } |S| = \min\big(\kappa,\, |C_{\mathrm{tgt}}|\big),
\end{aligned}
\end{equation}
with $\kappa = 3$ in every reported main experiment and $\tilde{X}$ the standardized
descriptors. A9 removes each rule in turn and adds a reference control of
3 episodes drawn at random from the cluster, with the target cluster fixed by the
memory in every arm so that only the composition of the set varies. All figures in this
paragraph are means over the 3 ablation settings against a full-system reference of
92.0, and they are collected in Table~\ref{tab:context}.

\begin{table}[t]
\centering
\caption{The composition and the size of the context set (studies A9 and A12), as the
mean cost in success-rate points over the 3 ablation settings against a
full-system reference of 92.0. The A12 citation-count comparison uses Qwen3-32B.}
\label{tab:context}
\begin{tabular}{lr}
\toprule
Arm & Cost (pts) \\
\midrule
\multicolumn{2}{l}{\emph{A9, composition; 3 citations}} \\
Drop the forced representative & $-3.2$ \\
Drop the diversity fill & $-3.2$ \\
Drop the worst-loss seed & $-3.27$ \\
3 episodes drawn at random & $-3.6$ \\
\midrule
\multicolumn{2}{l}{\emph{A12, number of citations; Qwen3-32B}} \\
Top 3 by plain peak-loss rank & $-2.13$ \\
2 citations & $-1.93$ \\
5 citations, Qwen3-32B & no gain \\
\bottomrule
\end{tabular}
\end{table}

The 3 single-rule removals have similar costs, showing that each contributes
comparably to the context-set construction. The forced representative guarantees the prescription model sees an
example of the mode it has been instructed to fix, without which the cited episodes can
all come from the cluster's periphery. The diversity fill keeps a context set of 3
episodes from all crowding near the loss peak, where it would describe the mode narrowly
and draw a narrow correction. The worst-loss seed sharpens the description of the mode.
Random selection of 3 episodes costs 3.6 points, more than any single-rule removal,
so the 3 rules together beat an unstructured draw and are complementary rather than
redundant.

The relative magnitudes place the context-set contribution in perspective. The spread from the full context set to
random selection is 3.6 points, larger than the constraint store and still less than the
clustering. A12 sharpens the point by varying the \emph{number} of cited episodes
jointly with the selection rule. At the same context size, the 3-rule construction
outperforms plain peak-loss ranking by 2.13 points and wins on each of the 3
settings. Citing only 2 episodes costs 1.93 points, while citing 5 provides no
measurable gain over 3 with Qwen3-32B. The cap of 3 therefore balances
contextual coverage with prompt length for the prescription model used in the main
experiments. Study A12b examines whether this choice transfers to a stronger alternative
model. Citation counts below 2 are not applicable to bridging, which requires a pair
of cited failures to define an intermediate placement.

\paragraph{A12b, prescription-model sensitivity to context-set size.}
The preceding A12 comparison was conducted with Qwen3-32B, the
prescription model used in the DISEIL experiments of Section~\ref{sec:study}. We additionally
tested whether the effect of the context-set cap depends on the prescription
model by comparing Qwen3-32B with Sonnet 5 at
$\kappa\in\{3,5\}$. The comparison was run on GridWorld (image) and
Door (image), with 3 seeds per arm. All other components of the
acquisition and training pipeline were held fixed.

Table~\ref{tab:model_context_interaction} shows a consistent crossover between
the 2 models. At $\kappa=3$, Qwen3-32B exceeds Sonnet 5 by $3.5$ points
on GridWorld (image) and $4.0$ points on Door (image). Increasing the
context-set size to $\kappa=5$ improves Sonnet 5 by $5.0$ and $6.5$ points,
respectively, after which it exceeds Qwen3-32B by $2.8$ and $3.1$ points.
In contrast, Qwen3-32B decreases by $1.3$ points on GridWorld (image) and
$0.6$ points on Door (image) when $\kappa$ is increased.

Averaged over the 2 settings, increasing $\kappa$ from 3 to 5
improves Sonnet 5 by $5.75$ points, whereas Qwen3-32B changes by
$-0.95$ points. The utility of additional cited episodes therefore depends
on the prescription model rather than only on prompt length. These results
support $\kappa=3$ for the Qwen3-32B instantiation used in the main
experiments, but they do not establish 3 episodes as a model-independent
optimum. Because the comparison covers 2 settings and 3 seeds, it is
interpreted as evidence of a model-by-context-set-size interaction rather
than as a general ranking of the 2 models.

\begin{table}[t]
\centering
\footnotesize
\caption{Prescription-model sensitivity to the context-set cap. Values are
final held-out success rate (per cent), reported as mean $\pm$ 1 standard
error over 3 seeds. The final column gives the within-Sonnet improvement
relative to $\kappa=3$ on the same task.}
\label{tab:model_context_interaction}
\setlength{\tabcolsep}{2pt}
\begin{tabular}{@{}>{\raggedright\arraybackslash}p{0.22\linewidth}cccc@{}}
\toprule
Task & $\kappa$ & Qwen3-32B & Sonnet 5 & \shortstack{Sonnet\\gain} \\
\midrule
GridWorld (image) & 3 & $91.5 \pm 0.3$ & $88.0 \pm 0.4$ & -- \\
GridWorld (image) & 5 & $90.2 \pm 0.4$ & $93.0 \pm 0.6$ & $+5.0$ \\
Door (image)      & 3 & $87.0 \pm 1.7$ & $83.0 \pm 1.2$ & -- \\
Door (image)      & 5 & $86.4 \pm 1.3$ & $89.5 \pm 1.3$ & $+6.5$ \\
\bottomrule
\end{tabular}
\end{table}

\paragraph{A11, the cluster count.} The cluster count is chosen per round by maximum
mean silhouette, which is standard practice, is used as such, and is not claimed as a
contribution \citep{rousseeuw1987silhouette, pedregosa2011sklearn}. A11 replaces the
adaptive choice with a fixed $k$. Silhouette selection wins on each of the 3
settings, and it beats the best fixed alternative, averaged over the 3, by 4.1
points: a fixed $k = 2$ costs 4.1 points against the adaptive rule, $k = 3$ costs 4.6,
$k = 4$ costs 7.4 and $k = 5$ costs 6.7. The effect is well outside the seed standard
error of the 3 settings, which runs from 0.6 to 1.6 points, so the adaptive rule is
defended by the size of its effect and not only by the consistency of its sign.

\begin{figure}[t]
\centering
\includegraphics[width=0.74\linewidth]{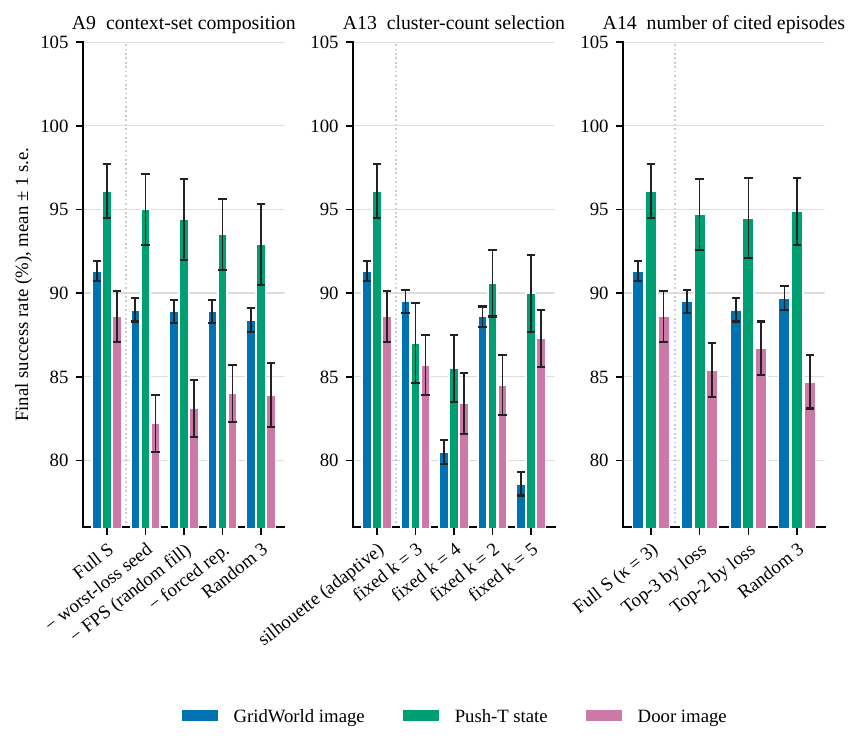}
\caption{3 ablations of the machinery inside individual steps, with arms ordered by
effect. Left: the composition of the context set (A9). Middle: silhouette-based
selection of the cluster count against fixed alternatives (A11). Right: the number of
cited episodes and the selection rule (A12). Bars are means over seeds with 1
standard error, on the 3 ablation settings.}
\label{fig:context}
\end{figure}

No single fixed $k$ is best across the 3 settings either: the best fixed value is
$k = 3$ on GridWorld (image), $k = 2$ on Push-T (state) and $k = 5$ on Door (image),
confirming the benefit of round-adaptive selection. Too few clusters merges distinct failure modes, so the memory
penalises a merged cluster and suppresses correction of a mode that was never addressed.
Too many splits a single mode across several clusters, so the memory cannot recognize that
the mode has been corrected and rotation is diluted across fragments of the same region.
The right number varies by setting and by round, and only the adaptive rule tracks it.
Figure~\ref{fig:context} plots A9, A11 and A12 together.

\section{Diagnostics, A13 to A16}

The final family of studies characterizes the discovered clusters, the use of bridging,
and the activity of the allocation machinery over the demonstration budget.

\paragraph{A13, semantic consistency of the geometric clusters.} A10 shows that the descriptor
produces well-separated clusters. A well-separated cluster need not correspond to a
single root cause, so A13 complements geometric silhouette with the consistency of the
system's root-cause labels. Purity runs from 0.84 to 0.91 across the 3 settings
and the number of distinct root causes per cluster runs from 1.35 to 1.86.
Table~\ref{tab:purity} gives the grid.

\begin{table}[t]
\centering
\footnotesize
\caption{Root-cause label purity per geometric cluster (study A13). Mean cluster purity
is the fraction of a geometric cluster's failures that share the dominant root cause
assigned by the prescription model. Mean root causes per cluster counts the distinct root
causes present in a cluster. Purity is scored against the prescription model's own labels,
so it measures internal agreement between geometric grouping and semantic labeling.}
\label{tab:purity}
\setlength{\tabcolsep}{3pt}
\begin{tabular}{@{}>{\raggedright\arraybackslash}p{0.38\linewidth}ccc@{}}
\toprule
Setting & Purity & Causes & Silhouette \\
\midrule
GridWorld 5x5 (image) & 0.89 & 1.62 & 0.58 \\
Push-T (state) & 0.91 & 1.35 & 0.64 \\
Door (image) & 0.84 & 1.86 & 0.56 \\
\bottomrule
\end{tabular}
\end{table}

The ordering is the informative part and it is the same ordering on both columns.
Push-T (state) has the best silhouette, 0.64, and the highest purity, 0.91. Door (image)
has the lowest silhouette, 0.56, the lowest purity, 0.84, and the most root causes per
cluster, 1.86. Geometric separation and semantic purity rise and fall together on these
3 settings: the descriptor variants that separate geometry most clearly also
produce the most semantically consistent clusters. A13 therefore provides complementary
evidence that the geometric partition aligns with the system's root-cause taxonomy.
Because the labels are assigned by the prescription model rather than by human
annotators, the measurement is interpreted as internal semantic consistency.

\paragraph{A14, the distribution of the cluster count.} A14 shows that the adaptive rule
uses the full candidate range. Pooled over the 308 clustered rounds of the 3 ablation
settings, $k = 3$ is the most frequently selected count at 26.3 percent, with $k = 4$
at 23.4, $k = 5$ at 21.4, $k = 2$ at 14.9 and $k = 6$ at 14.0. No value from 2 to 6
falls below 14 percent. Thus, the number of discovered failure modes varies by round
and is most often 3 or 4 rather than collapsing to a fixed value.

\begin{figure}[t]
\centering
\includegraphics[width=0.74\linewidth]{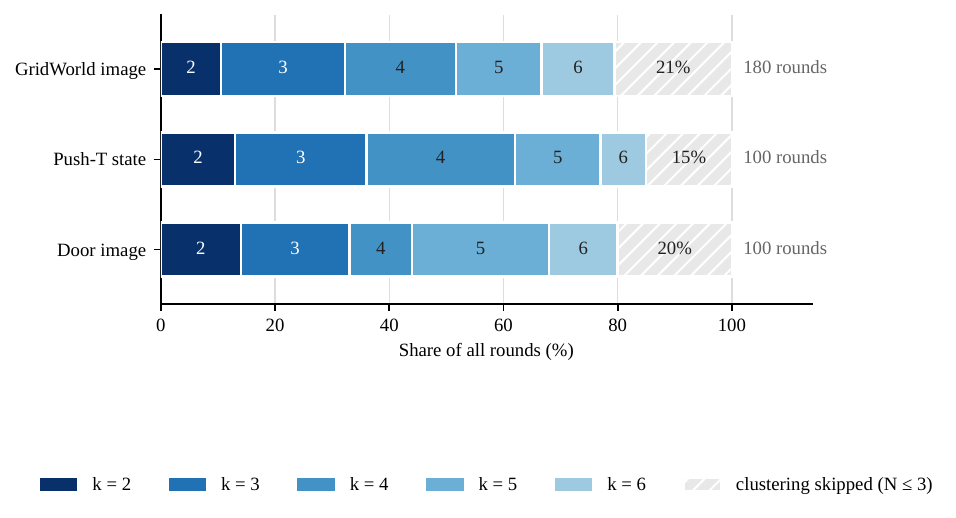}
\caption{Cluster count selected per round (study A14). Share of all rounds selecting
each cluster count $k$, with the rounds that skip clustering entirely, because fewer
than 4 failures remain, shown hatched.}
\label{fig:kdist}
\end{figure}

A14 also characterizes the intended small-set transition: 21 percent of
GridWorld (image) rounds, 15 percent of
Push-T (state) rounds and 20 percent of Door (image) rounds never cluster at all,
because fewer than 4 failures remain, and in those rounds each failure becomes its
own cluster and the budget is allocated by the fallback rule. Figure~\ref{fig:kdist}
shows the distribution with those rounds marked.

\paragraph{A15, the bridged share.} The share of accepted prescriptions that use
bridging is 24 percent on GridWorld (image), 28 percent on Push-T (state), and
21 percent on Door (image). Figure~\ref{fig:bridging} reports these values alongside
the A7 bridging knockout, linking the frequency of this prescription type to its
measured success-rate contribution.

\begin{figure}[t]
\centering
\includegraphics[width=0.74\linewidth]{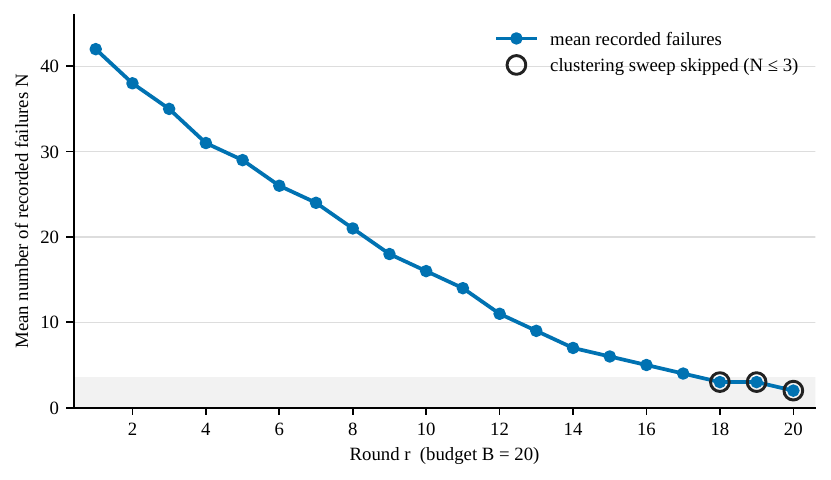}
\caption{Mean number of recorded failures per round over the budget (study A16). The
setting is Push-T (image), averaged over 5 seeds. Below 4 remaining failures the
clustering sweep is skipped and each failure becomes its own cluster; the shaded band
marks that region.}
\label{fig:failures}
\end{figure}

\paragraph{A16, failures per round.} Failures per round fall from 42 to 2 over
the budget, halving by round 8 and falling by an order of magnitude by round 17. The
decline shows that the system progressively reduces the failure set. On the instrumented
setting the descriptor, the clustering and the memory do their work through round 17,
and the last 3 rounds run the fallback rule on a handful of remaining failures. A14
confirms the pattern on the 3 ablation settings, where 15 to 21 percent of all
rounds skip clustering altogether. Figure~\ref{fig:failures} plots the curve.

This per-round diagnostic is instrumented on Push-T (image). A14 provides complementary
evidence across all 3 ablation settings by showing that 15 to 21 percent of rounds
reach the same small-failure-set regime and skip clustering.

A16 shows that DISEIL's allocation machinery is most active early on the instrumented
setting, which is consistent with the method concentrating its advantage in the
low-budget regime.

\clearpage
\section{Computational Cost, A17}

Study A17 directly measures the per-round computational cost of the reasoning pipeline,
complementing the success-rate results with an explicit account of the inference
trade-off.

\paragraph{The measurement.} Each of 5 settings runs a matched pair of runs, DISEIL
against SafeDAgger, on the same task, the same observation modality and the same
hardware, and the wall-clock and token cost of every round is read out of each run's own
telemetry. SafeDAgger is the baseline arm throughout. We report 2 protocols and
are interpreted separately. Protocol P1 reports the first round of each run. It is the only
protocol available in all 5 settings, and the first round holds the weakest policy
and therefore the most failures and the most model calls, so P1 is an upper bound on
steady-state cost rather than an average of it. Protocol P5 reports the mean and the
sample standard deviation over the language-model-active rounds of the runs that carried
a longer budget, matched arm for arm against the baseline's same-indexed rounds. The
spread quoted under P5 is the round-to-round spread inside a run. A P1 row and a P5 row
therefore describe distinct cost summaries.

\begin{table}[t]
\centering
\caption{Per-round wall-clock and token cost, DISEIL against SafeDAgger (study A17).
\emph{Shared} is the part of the round that both arms pay: a from-scratch policy retrain
and a held-out evaluation. \emph{Add-on} is every DISEIL-specific stage of the round,
which is the failure-screening rollout together with clustering, the vision-language
call, the reasoning call, the prescription and the feasibility check, less the
query-gate rollout the baseline runs in its place. \emph{Tokens} is the sum of the
vision-language and language-model tokens drawn in the round. Token counts are not
comparable across rows, because the settings were served by different backends and the
hidden reasoning tokens are read directly from the usage record on some and recovered
from the billed completion length on others; within-row wall-clock comparisons use
matched hardware and protocols.}
\label{tab:cost}
\footnotesize
\setlength{\tabcolsep}{4pt}
\begin{tabular}{llrrrrr}
\toprule
Protocol & Setting & Baseline (s) & DISEIL (s) & Shared (s) & Add-on (s) & Tokens \\
\midrule
P1 & Door (state) & 737.0 & 1{,}054.0 & 783.0 & $+270.0$ & 11{,}511 \\
P1 & Door (image) & 1{,}247.0 & 1{,}474.0 & 1{,}180.0 & $+293.0$ & 11{,}379 \\
P1 & Wipe (image) & 1{,}468.0 & 2{,}195.0 & 1{,}491.0 & $+700.0$ & 9{,}560 \\
P1 & Push-T (image) & 688.0 & 1{,}891.0 & 652.5 & $+1{,}232.1$ & 12{,}116 \\
P1 & GridWorld (image) & 54.6 & 118.0 & 51.1 & $+62.6$ & 9{,}735 \\
\midrule
P5 & Door (state) & 532.6 $\pm$ 145.5 & 782.6 $\pm$ 183.5 & 547.8 & $+232.8$ & 10{,}929 \\
P5 & Door (image) & 1{,}034.2 $\pm$ 150.2 & 1{,}168.8 $\pm$ 194.8 & 901.6 & $+266.2$ & 10{,}787 \\
P5 & Wipe (image) & 1{,}303.3 $\pm$ 144.1 & 1{,}991.0 $\pm$ 179.2 & 1{,}332.7 & $+655.7$ & 10{,}118 \\
\bottomrule
\end{tabular}
\end{table}

\paragraph{What the measurement says.} A round's wall-clock is dominated by the
from-scratch policy retrain and the held-out evaluation, and both arms pay all of it.
Under P1 that shared cost is 783 to 1,491 seconds per round on the 3 RoboSuite
settings, against a reasoning add-on of 270 to 700 seconds, and under P5 it is 548 to
1,333 seconds against an add-on of 233 to 656. The ratio of a DISEIL round to a baseline
round includes a large shared denominator: across the 2 protocols, the ratio ranges
from 1.13 to 2.75. The DISEIL-specific add-on ranges from 63 seconds per round on
GridWorld to 1,232 seconds on Push-T, with 9,560 to 12,116 reported tokens per round.
Together, the shared and add-on columns separate the cost of the underlying training
loop from the cost of DISEIL's selection pipeline.

Push-T has the largest measured add-on. Its shared cost is the second
smallest in the sweep and its baseline arm is unusually short, because SafeDAgger's loop
runs until the intervention budget is spent and therefore halts after its first
intervened episode, which is 1 episode and 6.3 seconds, while DISEIL screens a fixed
60. Push-T also issues 20 language-model calls, compared with 7 on Door.
GridWorld has the smallest wall-clock add-on, while still drawing 9,735 tokens; its
constraint block accounts for 54 percent of the prompt budget, the largest share among
the measured settings.

\paragraph{What this means for the framework.} The language and vision-language models
run only at demonstration-selection time. They are never in the control loop and they do
not run at execution, so their cost is paid once per round, and on the 3 RoboSuite
settings it is amortized over a retrain and an evaluation that together cost more than
the reasoning does. The resource being traded is model inference against expert
demonstration time, and on the settings measured here a round buys 1 demonstration for
an extra 63 to 1,232 seconds of inference. Whether that trade is worth making depends on
what a demonstration costs, which for a scripted or oracle expert is negligible but for a
human demonstrator is a person's time. DISEIL is therefore most attractive when expert
supervision is scarce relative to selection-time compute. A4 additionally provides a
reduced-compute operating point: replacing the reasoning stack with the deterministic
cluster-representative heuristic costs 1.33 success-rate points on average while
retaining performance above the evaluated baselines.

\clearpage
\section{Prescription Confidence}

At the moment it issues a prescription, the prescription model also emits an integer
confidence between 0 and 100 together with a one-line rationale, reporting how likely it
believes the resulting demonstration is to improve the policy. That number was added to
the prompt as an instrument, to find out whether the model has any usable forecast of
the value of the round it is about to spend. It is scored against $\Delta$SR, the change
in the policy's success rate on the round-level rollout evaluation across that round.

\begin{figure}[t]
\centering
\includegraphics[width=0.74\linewidth]{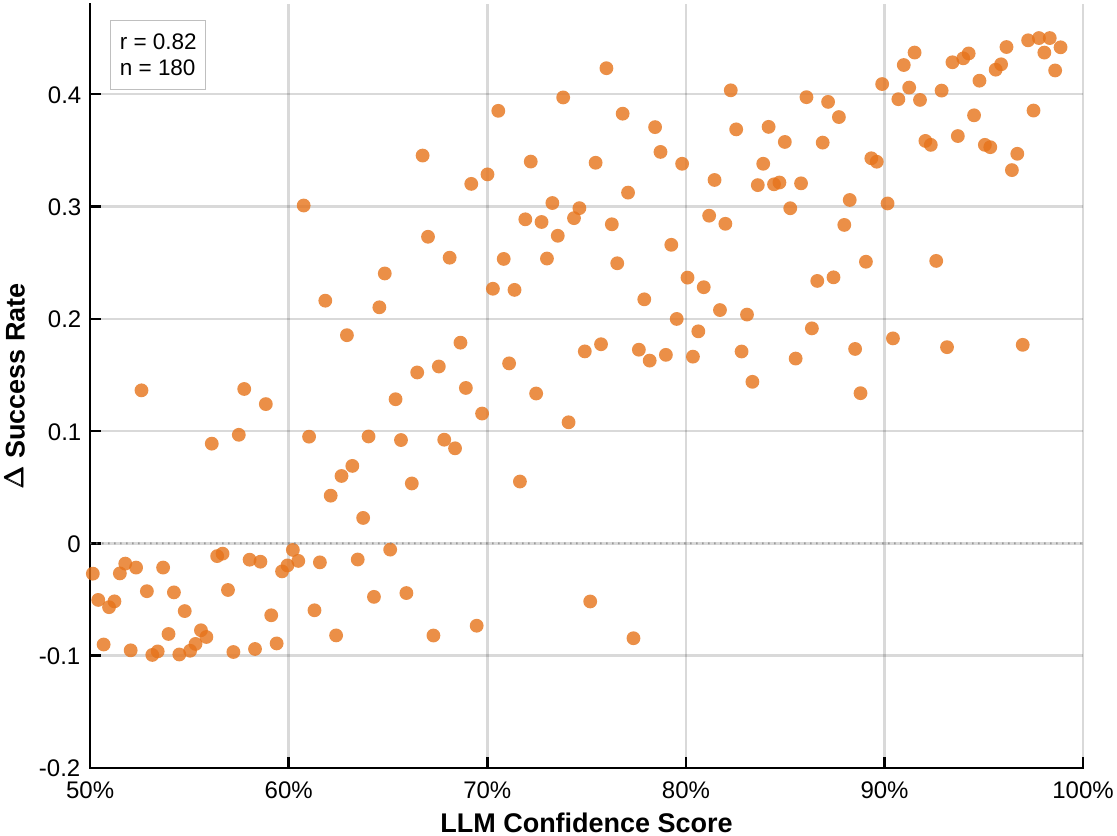}
\caption{Reported prescription confidence against realized improvement. The confidence
the prescription model reports at prescription time, against the change in the policy's
success rate on the round-level rollout evaluation that the resulting demonstration
produced. GridWorld (image), 180 prescriptions, Pearson $r = 0.82$.}
\label{fig:confidence}
\end{figure}

The Pearson correlation between the reported confidence and the realized $\Delta$SR runs
from 0.82 to 0.89 across the 10 settings. Quoting each task as state then image:
GridWorld 0.88 and 0.82, Push-T 0.87 and 0.88, Lift 0.88 and 0.89, Wipe 0.82 and 0.86,
Door 0.83 and 0.82. Figure~\ref{fig:confidence} shows the GridWorld (image) scatter,
where $r = 0.82$ over 180 prescriptions, the full count that 9 seeds at a budget of
20 supply.

The temporal ordering makes this a prospective rather than retrospective signal. The confidence is reported blind, at
prescription time. At that moment the demonstration has not been collected, the expert
has not been called, the policy has not been retrained, and the re-rollout that produces
$\Delta$SR has not been run. The success-rate signal arrives only after all 3 of
those steps have completed, by which point the round's unit of budget has already been
spent. The model therefore reports confidence before observing the outcome, and the correlations of $0.82$ to $0.89$ show a strong positive association between reported confidence and subsequent policy improvement.

The correlation is measured on DISEIL runs, and confidence is recorded as a diagnostic:
no demonstration is skipped, deferred, or re-prescribed on the basis of the score.
Accordingly, the result establishes confidence as a strongly correlated prospective signal
within the reported DISEIL runs without attributing any of the policy improvement to
confidence-based gating.

\section{DISEIL Prompt Templates}

\subsection{Prompts and the Failure Taxonomy}

Every run stores its exact prompts and replies. The 3 model calls are the perception
call, the reasoning call and the prescription call, and their templates are reproduced
below as text. The anchor $t^\star$ is the first uncertain step, the first step at which
the policy's per-step loss stays above threshold for $K$ consecutive steps; it is
distinct from peak loss, which is used only to rank a mode's severity.

The perception call takes a system instruction and 3 frames of a cited failure: the
start of the episode, the first uncertain step $t^\star$, and the final step.

\begin{figure}[H]
\centering
\begin{minipage}{\linewidth}
{\footnotesize
\begin{verbatim}
SYSTEM:  You are analyzing a robot manipulation failure from rendered frames. Be
         concrete and spatial; describe what you actually see, not generic advice.

USER:    You are analyzing a robot manipulation failure. The attached frames are, in
         order: start, first_uncertain, end (the first_uncertain frame is the
         policy's first uncertain step t*, where per-step loss first stays high).
         Task: {task_description}
         Describe what went wrong. Focus on: where in the trajectory the failure
         occurs, the robot/gripper configuration at t*, and what object or
         contact state caused it. ~120 words, concrete and spatial.
         [start frame] [first_uncertain frame] [end frame]
\end{verbatim}
}
\end{minipage}
\end{figure}

The reasoning call is constrained to the vocabulary of the task's constraint store and
to strict JSON. The closed taxonomy is visible in the prompt: the root cause is one of
6 categories and the phase one of 5, and the model's job is assignment rather than
invention.

\begin{figure}[H]
\centering
\begin{minipage}{\linewidth}
{\footnotesize
\begin{verbatim}
SYSTEM:  You are a robot-manipulation failure analyst. Classify the root cause and
         trajectory phase using ONLY the provided categories and the KAG facts.
         Output strict JSON, no prose, no code fences.

USER:    TASK: {task_description}
         {kag_text}
         VLM FAILURE DESCRIPTION (the only visual evidence): {vlm_report}
         Identify the root cause category and the trajectory phase where the failure
         occurred.
         root_cause in [grasp_failure, approach_failure, placement_error,
                        contact_instability, pose_mismatch, timeout]
         phase      in [pre_grasp, grasp, transport, placement, insertion]
         Output ONLY this JSON:
         {"root_cause": "<one of the categories>",
          "phase": "<one of the phases>",
          "rationale": "<one sentence grounded in the VLM description
                         and a KAG fact>"}
\end{verbatim}
}
\end{minipage}
\end{figure}

The 2 closed vocabularies are read from the task's store of environmental constraints,
so a task that has no insertion phase does not offer one, and the rationale field forces
the label to be traceable to both the visual evidence and a named constraint.

The prescription call states the budget rule to the model, offers the 2 arms of the
prescription, and requires a confidence line.

\begin{figure}[H]
\centering
\begin{minipage}{\linewidth}
{\footnotesize
\begin{verbatim}
SYSTEM:  You are a demonstration coach for an interactive imitation-learning
         loop. The target failure mode has already been selected by the
         geometric allocation stage. You decide only HOW to spend ONE expert
         demonstration intended to address this selected mode.

         Do not re-cluster the failures, select another failure mode, or
         generate policy actions. Ground the request only in the supplied
         failure evidence and environmental constraints.

         Reason briefly, then end with EXACTLY two lines:
         (1) a decision line in the required format; and
         (2) a confidence line:
         CONFIDENCE: <integer 0-100> - <one-line rationale> reporting how
         confident you are that this demonstration will improve the policy.

USER:    TASK: {task_description}
         PRESELECTED TARGET FAILURE MODE: {cluster_summary}
         CITED FAILURES FROM THIS MODE: {failure_analyses}
         FEEDBACK: {validation_feedback}

         AVAILABLE PRESCRIPTION TYPES:

         (A) SELECT ep<ID> - one cited failure represents the selected mode.
             Restore that failure's anchored state and ask the expert to
             continue from the first uncertain step t*. Use SELECT when the
             mode is geometrically tight or one cited failure provides a
             suitable representative correction.

         (B) BRIDGE ep<ID>,ep<ID> - no single cited failure adequately
             represents the selected mode. Propose ONE new, feasible starting
             configuration in an intermediate region between two or three
             cited failures. Use BRIDGE when the failures are geometrically
             spread but represent the same selected mode.
\end{verbatim}
}
\end{minipage}
\end{figure}

\subsection{Environmental Constraints: Representative Examples}

The store of environmental constraints for a task is a JSON document with a fixed
schema: metadata, typed nodes with key-value properties, relations between them, and a
block of reasoning implications, 1 per failure mode plus a workspace constraint and a
non-emptiness rule. A renderer turns the document into the text block injected into the
reasoning and prescription prompts. The store is authored once per task, and its
constraints are measurements of the environment rather than opinions about it.

Push-T stores its bounds as typed workspace nodes and its controller as a node in its
own right.

{\footnotesize
\begin{verbatim}
{"id":"ws_tee","type":"Workspace",
 "label":"Reliable tee init range",
 "properties":{"x":[-0.20,0.20],
               "y":[-0.25,0.05],
               "z":0.021}},
{"id":"ws_tcp","type":"Workspace",
 "label":"Reliable tcp range",
 "properties":{"x":[-0.35,0.35],
               "y":[-0.35,0.35],
               "z":[0.02,0.08]}},
{"id":"ctrl","type":"Controller",
 "label":"pd_joint_pos /
           rel_joint_pos",
 "properties":{
   "policy_action":"7 joint deltas
                    (rel_joint_pos)",
   "expert_action":"PPO ->
        joint_delta_pos (same
        7-joint space)"}}
\end{verbatim}
}

The predicate the feasibility loop checks is stored as an implication and is written in
the imperative, because it is addressed to the model as much as to the checker.

\begin{quote}
\raggedright
\texttt{"workspace\_constraint": "Every prescribed config MUST keep tee\_xyz within
x[-0.20,0.20] y[-0.25,0.05] z=0.021 and tcp\_xyz within x[-0.35,0.35] y[-0.35,0.35]
z[0.02,0.08]; out-of-range poses are dropped (the PPO expert is unreliable there) and
waste the round."}
\end{quote}

That constraint is also where the Push-T expert's partial competence is encoded. The
expert pushes in a single rotational direction only, so the configurations it cannot
demonstrate are excluded before a prescription is issued rather than discovered after
the expert has failed on 1.

Door's constraint is tighter by an order of magnitude, and the numbers are a padded
empirical measurement of the environment's own reset sampler, so that a prescribed
configuration cannot leave the task's native reset distribution.

{\footnotesize
\begin{verbatim}
{"id": "ws_door", "type": "Workspace",
 "label": "Reliable door-frame range",
 "properties": {"x": [-0.135, -0.108],
                "y": [-0.366, -0.340],
                "z": 1.10,
                "yaw_rad":
                    [-1.82, -1.57]}},
{"id": "succ",
 "type": "SuccessCondition",
 "label": "Door open",
 "properties": {
   "metric": "hinge_qpos > 0.3 rad",
   "info_key": "success"}}
\end{verbatim}
}

For a discrete task the environmental constraint is not a bounding box but a
reachability predicate, and GridWorld's store states it as one. A prescribed layout must
place the start cell, the goal cell and the 3 obstacle cells as distinct in-grid
cells, with a minimum separation between start and goal and a breadth-first path from
one to the other that avoids the obstacles. A layout that fails the predicate is
rejected before it reaches the expert.

Beyond rejecting infeasible layouts, the store also fixes which prescription arms a task
offers at all: where a task's constraints declare bridging inapplicable, the bridging arm
is omitted from the prompt and only targeted selection remains. Both jobs, constraining
where a demonstration may be placed and fixing which prescription arms exist, are done by
knowledge that is written down and checkable, which is the sense in which the framework's
model of the environment is explicit rather than implicit in a network's weights.

\subsection{Qualitative Examples and Operational Characteristics}

\paragraph{The modes the partition discovers.} Figure~\ref{fig:failure_modes} shows the 3
failure modes discovered on Push-T (image) with 3 sampled members of each. The modes
are behaviorally distinct: the block is brought to the goal region but left almost fully
inverted; the arm never establishes a working contact; the block is pushed but abandoned
at a moderate orientation error and far from the end-effector. The clusters are found
from geometry alone and the naming pipeline supplies their labels from the task store's
own vocabulary. The value 3 is not the number of failure modes on these tasks. It is the count
most often selected when the silhouette criterion runs, and the count varies by round, as
study A14 reports.

\paragraph{A round, end to end.} The cited failures are handed to the prescription model
1 line each and the model answers in 2 lines. The round below is taken verbatim from
a Wipe log.

{\footnotesize
\begin{verbatim}
DOMINANT FAILURE CLUSTER
(members with their VLM+analysis
findings):
  - ep3000001:
      object_xy=(0.221,-0.027)
      progress=74/500
      peak_loss=0.0434
      root_cause=contact_instability
      phase=transport
  - ep3000003:
      object_xy=(0.165,-0.068)
      progress=146/500
      peak_loss=0.0181
      root_cause=contact_instability
      phase=transport
  - ep3000006:
      object_xy=(0.123,-0.157)
      progress=234/500
      peak_loss=0.0186
      root_cause=pose_mismatch
      phase=placement
===== RESPONSE =====
SELECT ep3000001
CONFIDENCE: 85 - Ep3000001's
contact_instability root cause (higher
peak_loss) at early progress best
represents the dominant failure mode,
ensuring the expert demonstration
directly addresses unstable wiping
pressure causing missed coverage.
\end{verbatim}
}

\paragraph{A mixed-label cluster.} The same record illustrates how majority labeling
operates when a geometric cluster contains more than 1 semantic label. Of the 3 cited members, 2 carry the label
\texttt{contact\_instability} and the third carries \texttt{pose\_mismatch}, so this mode
has a purity of 2 thirds. The purity measured across the ablation settings, 0.84 to
0.91 in study A13, summarizes this behavior across clusters. In this example, the
prescription selects an episode carrying the majority label and targets the dominant
contact-instability cause.

\paragraph{Fallback and small-set behavior.} The framework includes explicit handling
for 2 operational edge cases. If no feasible prescription is produced within
$J_{\max}$ attempts, the nearest untried failure supplies the demonstration start; A6
shows that this fallback is used in 27 to 35 percent of rounds when the constraint store
is removed. When fewer than 4 failures remain, each failure becomes a singleton and
the same deterministic rule allocates the round; A14 and A16 show that this transition
occurs in 15 to 21 percent of rounds and is concentrated near the end of the budget.
These mechanisms provide a deterministic demonstration request whenever a recorded
failure remains, even when the full clustering or prescription path is unavailable.

\section{Scope of the Evidence}

\paragraph{Experimental scope.} The reported evaluation uses 5 simulated tasks under
state and image observations. Experts are human, scripted, or learned as specified in
Appendix~B, and every comparison uses the same per-task expert and demonstration budget.
The results therefore establish supervision efficiency under controlled, matched expert
conditions; interactive human teaching on the robot tasks is outside the reported
scope.

\paragraph{Representation and ablation scope.} DISEIL uses the task-specific,
6-dimensional geometric descriptors in Table~\ref{tab:descriptor}. A10 validates the
descriptor width through silhouette, and A13 measures agreement with root-cause labels
assigned by the prescription model. The selector's cross-round history is the geometric
cluster memory; it does not index the raw contents of the aggregated training set. A3
removes the descriptor and memory together with clustering and therefore quantifies the
integrated allocation stack, as stated in Appendix~F.

\paragraph{Temporal and computational scope.} The per-round failure-count curve in A16
is instrumented on Push-T (image), while A14 measures the small-set transition across all
3 ablation settings. The language and vision-language models run only during
demonstration selection, never during policy execution. A17 reports their measured
selection-time overhead under the matched P1 and P5 protocols.

\paragraph{Prescription-model scope.} The DISEIL results of Section~\ref{sec:study} and the original A12
context-set-size comparison use Qwen3-32B as the prescription model. A12b shows a
different response to additional cited episodes for Sonnet 5 on GridWorld (image) and
Door (image). The evidence therefore supports $\kappa=3$ for the reported Qwen3-32B
instantiation, not as a model-independent optimum; the cross-model comparison is a
sensitivity result based on 2 settings and 3 seeds per arm.

\end{document}